\documentclass{article}

\usepackage{microtype}
\usepackage{graphicx}
\usepackage{subfigure}
\usepackage{booktabs} %

\usepackage{hyperref}

\usepackage[accepted]{icml2024}

\usepackage{amsmath}
\usepackage{amssymb}
\usepackage{mathtools}
\usepackage{amsthm}

\usepackage[capitalize,noabbrev]{cleveref}
\usepackage{multirow}
\usepackage{graphicx}
\usepackage{tabu}
\usepackage{subcaption}
\usepackage[ruled,noend,algo2e]{algorithm2e}

\theoremstyle{plain}

\theoremstyle{definition}

\theoremstyle{remark}

\DeclareMathOperator*{\argmax}{arg\,max}

\usepackage[textsize=tiny]{todonotes}

\icmltitlerunning{Neural NeRF Compression}

\begin{document}

\twocolumn[
\icmltitle{Neural NeRF Compression}

\icmlsetsymbol{equal}{*}

\begin{icmlauthorlist}
\icmlauthor{Tuan Pham}{uci}
\icmlauthor{Stephan Mandt}{uci}
\end{icmlauthorlist}

\icmlaffiliation{uci}{Department of Computer Science, University of California Irvine}

\icmlcorrespondingauthor{Tuan Pham}{tuan.pham@uci.edu}

\icmlkeywords{Machine Learning, ICML}

\vskip 0.3in
]

\printAffiliationsAndNotice{}  %

\begin{abstract}
Neural Radiance Fields (NeRFs) have emerged as powerful tools for capturing detailed 3D scenes through continuous volumetric representations. Recent NeRFs utilize feature grids to improve rendering quality and speed; however, these representations introduce significant storage overhead. This paper presents a novel method for efficiently compressing a grid-based NeRF model, addressing the storage overhead concern. Our approach is based on the non-linear transform coding paradigm, employing neural compression for compressing the model's feature grids. Due to the lack of training data involving many i.i.d scenes, we design an encoder-free, end-to-end optimized approach for individual scenes, using lightweight decoders. 
To leverage the spatial inhomogeneity of the latent feature grids, we introduce an importance-weighted rate-distortion objective and a sparse entropy model employing a masking mechanism. Our experimental results validate that our proposed method surpasses existing works in terms of grid-based NeRF compression efficacy and reconstruction quality. 
\end{abstract}

\section{Introduction}
Over the past few years, the field of 3D scene modeling and reconstruction has been revolutionized by the advent of Neural Radiance Fields (NeRF)  methodologies \citep{mildenhall2021nerf, zhang2020nerf++, barron2021mip}. NeRFs offer a sophisticated method for 3D reconstruction, with the ability to synthesize novel viewpoints from limited 2D data. Yet, the original NeRF model requires millions of MLP queries, which causes slow training and rendering. 

To address these efficiency concerns, recent NeRF advancements have transitioned to the integration of an explicit grid representation \citep{yu2021plenoctrees, sun2022direct, fridovich2022plenoxels, chen2022tensorf, fridovich2023k, chan2022efficient}. While significantly accelerating training and rendering processes, this change also poses a new challenge: the storage cost for saving the explicit grid NeRF representation increases. This problem is crucial, especially in real-world (e.g., large-scale VR and AR) applications where storage and transmission impose critical constraints.

Our work seeks to significantly reduce the storage costs of NeRFs. Inspired by neural image compression methodology~\citep{yang2023introduction}, we apply non-linear transform coding techniques \citep{balle2020nonlinear} to compress the explicit grid NeRF representation efficiently. However, we sidestep the conventional auto-encoder approach in favor of an iterative inference framework, in which we jointly optimize the latent code along with a lightweight decoder. We further take account of the NeRF grid importance scores while reconstructing the scene to boost the efficiency of our compression model. Lastly, we propose a novel entropy model that masks uninformative feature grid points. Utilizing a rate-distortion objective, we can choose from various compression levels. Our proposed approach departs from previous works on compressing explicit grid NeRF representations \citep{li2023compressing, li2023compact, deng2023compressing} based on voxel pruning and/or vector quantization \citep{gray1984vector} while taking into account the varying importance levels of different voxel grid locations. 

To show the effectiveness of our proposed method, we perform extensive experiments on four different datasets. Our results show that our model is capable of compressing diverse NeRF scenes to a much smaller size and improves over previous works in terms of rate-distortion performance.

\section{Background}
\subsection{Neural Radiance Fields} \label{sec2.1}
Neural radiance fields \citep{mildenhall2021nerf} mark a paradigm shift in 3D scene representation using deep neural networks. Unlike traditional approaches that employ discrete structures such as point clouds or meshes, NeRFs model a scene using a continuous volumetric function $F: (\mathbf{x}, \mathbf{d}) \rightarrow (\mathbf{c}, \sigma)$. Here, an input comprising of spatial coordinates $\mathbf{x}$ and a viewing direction $\mathbf{d}$ is mapped to an output representing color $\mathbf{c}$ and volume density $\sigma$. 

For each pixel, the estimated color \(\hat{C}(\mathbf{r})\) for the corresponding ray \(\mathbf{r}\) can be calculated by:
\begin{equation}
\label{eq:rendering}
\begin{aligned}
\hat{C}(\mathbf{r}) = \sum_{i=1}^N T_i \cdot \alpha_i \cdot \mathbf{c}_i,
\end{aligned}
\end{equation}
with the following definitions:
\begin{itemize}
    \item \(\alpha_i = 1 - \exp(-\sigma_i \delta_i)\) is the probability of light being absorbed at the \(i\)-th point, dependent on the volume density \(\sigma_i\) and the distance \(\delta_i\) between adjacent sampled points on the ray.
    \item \(T_i = \prod_{j=1}^i (1 - \alpha_j)\) represents the accumulated transmittance, or the remaining light that has not been absorbed before reaching the \(i\)-th point.
\end{itemize}

NeRF is then trained to minimize total squared error loss  between the rendered and true pixel colors.
\begin{equation}
    \mathcal{L}_{render} = \sum_\mathbf{r} || \hat{C}(\mathbf{r}) - C(\mathbf{r}) ||_2^2
\end{equation}
Despite NeRF's ability to provide intricate scene details with a relatively compact neural network, the computational demand remains a significant constraint. The evaluation over the volume often requires thousands of network evaluations per pixel. To reduce training and inference time, recent research has employed explicit grid structure into NeRF. More specifically, they introduce voxel grids \citep{sun2022direct, fridovich2022plenoxels} or decomposed feature planes \citep{chen2022tensorf, fridovich2023k, chan2022efficient} into the model, and query point features via trilinear or bilinear interpolation. While this notably speeds up training and inference, it does come at the expense of greater storage needs from saving the feature grids.

\subsection{Neural Compression}
Neural compression utilizes neural networks to perform end-to-end learned data compression~\citep{yang2023introduction}. Traditional compression algorithms are handcrafted and specifically tailored to the characteristics of the data they compress, such as JPEG \citep{wallace1991jpeg} for images or MP3 for audio. In contrast, neural compression seeks to learn efficient data representations directly from the data, exemplified by the nonlinear transform coding paradigm \citep{balle2020nonlinear}.

Existing lossy neural compression methods \citep{balle2016end, balle2018variational, minnen2018joint, cheng2020image, matsubara2022supervised, yang2023lossy, yang2023computationally, yang2023insights} often leverage an auto-encoder architecture \citep{kingma2013auto}. Here, an encoder $E$ maps data $\mathbf{X}$ to continuous latent representations $\mathbf{Z} = E(\mathbf{X})$. This continuous  $\mathbf{Z}$ is then quantized to integers by $Q$, resulting in  $\hat{\mathbf{Z}} = Q(\mathbf{Z})$. An entropy model $P$ is used to transmit $\hat{\mathbf{Z}}$ losslessly. Finally, a decoder $D$ receives the quantized latent code $\hat{\mathbf{Z}}$ and reconstructs the original data $\hat{\mathbf{X}} = D(\hat{\mathbf{Z}})$. One commonly trains the encoder $E$, the decoder $D$ and the entropy model $P$ jointly using a rate-distortion objective:
\begin{equation}
\begin{aligned}
\mathcal{L}(E, D, P) = &\mathbb{E}_{\mathbf{X} \sim p(\mathbf{X})} [d(\mathbf{X}, D(Q(E(\mathbf{X}))) \\
&- \lambda \log_2 P(Q(E(\mathbf{X})))]
\end{aligned}
\end{equation}
where $d(\cdot, \cdot)$ is a distortion loss and the second term is the rate loss that measures the expected code length. The parameter  $\lambda$ balances between the two loss terms. At training time, the quantizer $Q$ is typically replaced with injecting uniform noise \citep{balle2016end}. See \citep{yang2023introduction} for a detailed review of neural compression.

\section{Method}
In this section, we describe our method for grid-based NeRF compression. Our primary focus is on the compression of the TensoRF-VM model \citep{chen2022tensorf}, characterized by its decomposed 2D feature plane structure \citep{kolda2009tensor}. We select TensoRF-VM because of its proficient 3D scene modeling capabilities, often outperforming alternative methods like Plenoxels \citep{fridovich2022plenoxels} or DVGO \citep{sun2022direct}. Our method has the potential to be applied to other grid-based NeRF architectures.

\paragraph{Problem setting.}  We have a TensoRF-VM model that was pre-trained for a single scene, and our task is to reduce its size through compression while maintaining its reconstruction quality. We assume that we have access to the training dataset comprising view images at compressing time.

\paragraph{Notation.} The three feature planes (or matrix components) of TensoRF-VM are denoted by $\{\mathbf{P}_i\}_{i=1}^3$, in which subscript $i$ signifies the index of the planes and each $\mathbf{P}_i \in \mathbb{R} ^ {C_i \times H_i \times W_i}$. 
In practice, $\{\mathbf{P}_i\}_{i=1}^3$ is the channel-wise concatenation of the density planes and appearance planes of TensoRF-VM.
The vector components are not considered in our compression and, hence, are not represented in our notation. For indexing a specific spatial grid location $j$ in the feature plane $i$, we employ a superscript, represented as $\mathbf{P}_i^j$.

\subsection{Compressing the feature planes} \label{sec3.1}
\begin{figure*}[!t]
\begin{center}
\includegraphics[width=\textwidth]{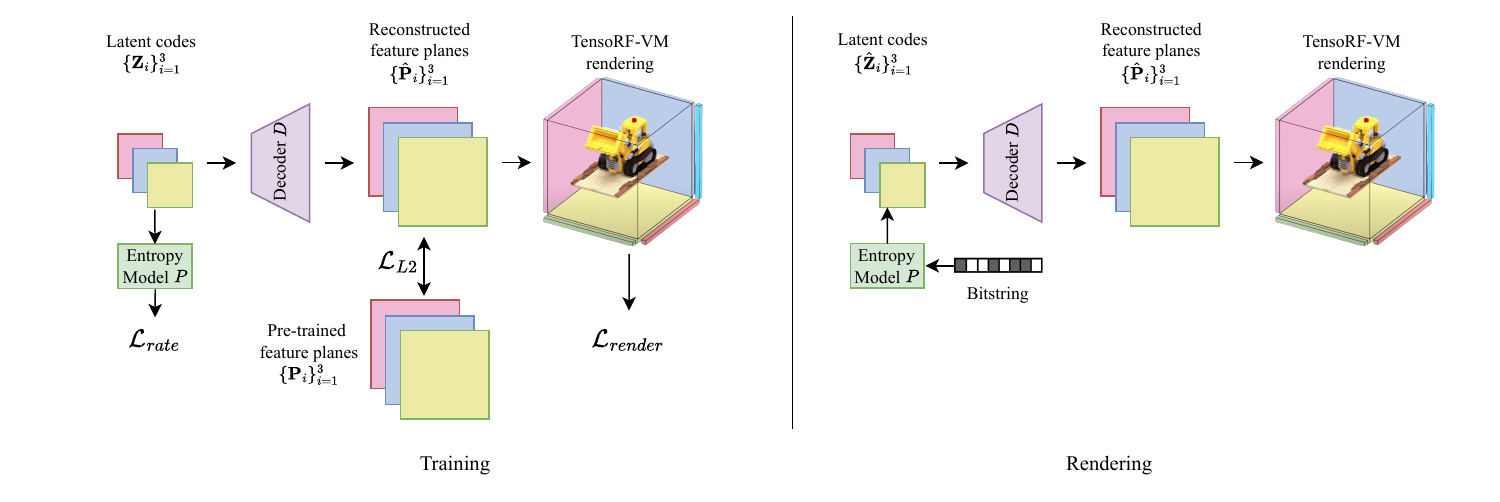}
\end{center}
\vspace{-5mm}
\caption{\textbf{Overview of our model.} At training time (left), we learn the three latent codes $\{\mathbf{Z}_i\}_{i=1}^3$ to reconstruct the three frozen feature planes $\{\mathbf{P}_i\}_{i=1}^3$. The reconstructed feature planes $\{\hat{\mathbf{P}}_i\}_{i=1}^3$. are used to render the scene and calculate the rendering loss. The entropy model $P$ is used to calculate the rate loss and compress the latent codes to bitstring. At rendering time (right), we use $P$ to decompress the bitstring to latent codes $\{\hat{\mathbf{Z}}_i\}_{i=1}^3$ and then reconstruct the feature planes $\{\hat{\mathbf{P}}_i\}_{i=1}^3$.}
\label{fig:model}
\end{figure*}

Most storage for compressing TensoRF-VM is spent on the feature grids. To illustrate this, we analyze a trained model for the Lego scene from the Synthetic-NeRF dataset \citep{mildenhall2021nerf}. In this model, the 2D feature planes take $67.61$ MB, while the other components, such as the rendering MLP, the rendering mask, and the vector components, take only $1.21$ MB. Given this disparity, we focus on compressing TensoRF-VM's feature planes. 

In more detail, we can define an encoder $E$  that embeds the three feature planes $\{\mathbf{P}_i\}_{i=1}^3$ to latent codes $\{\mathbf{Z}_i\}_{i=1}^3$, in which the $\mathbf{Z}_i$ may have lower resolution than $\mathbf{P}_i$. The latent codes are quantized to $\{\hat{\mathbf{Z}}_i\}_{i=1}^3$ and compressed with entropy coding using an entropy model $P$. At rendering time, we decompress the quantized latent codes $\{\hat{\mathbf{Z}}_i\}_{i=1}^3$ and forward them to the decoder $D$ to reconstruct the three feature planes $\{\hat{\mathbf{P}}_i\}_{i=1}^3$. We then use $\{\hat{\mathbf{P}}_i\}_{i=1}^3$ to query sampling point features and render the scene. The compressed NeRF model includes the compressed latent codes, the decoder, the entropy model, and the other components. 

It is crucial to highlight that we only need to reconstruct the three feature planes once, and all subsequent querying operations for the sampling points are executed on these reconstructed planes. Thus, the decompression process only adds minimal overhead to the overall rendering procedure.

\paragraph{Per-scene optimization.} The conventional approach to neural image compression \citep{balle2016end, balle2018variational} involves training a compression model on a big dataset containing thousands of images. However, applying this same training method to NeRF compression presents three challenges: First, we need a dataset with numerous 3D objects. Although datasets like Objaverse \citep{deitke2023objaverse} or Objaverse-XL \citep{deitke2024objaverse} exist, they are synthetic datasets and only contain a single object for each scene. Additionally, pre-trained NeRF models are required for every 3D object in the dataset, demanding significant computational resources and storage. Finally, we cannot adapt other components of the NeRF model, such as the rendering MLP and the vector components of the TensorF-VM model. Due to these challenges, we optimize each NeRF scene individually, a process we refer to as \textbf{per-scene optimization}. In this approach, the compressor is overfitted to each NeRF scene, which results in improved compression performance.

\vspace{-2mm}

\paragraph{Transform coding without encoder.} In nonlinear transform coding~\citep{balle2020nonlinear, yang2023introduction}, one usually employs an encoder to obtain the latent code of a new data point via amortized inference \citep{kingma2013auto, gershman2014amortized}. This is essential for compressing a new data point quickly in a single network pass. Nonetheless, in the case of per-scene TensoRF-VM compression, our primary objective is to compress merely the three feature planes, and our decoder is overfitted to a single scene. Moreover, using an encoder for amortized inference leads to an irreducible amortization gap in optimization \citep{cremer2018inference, marino2018iterative}, which has been shown to degrade compression performance \citep{campos2019content, yang2020improving}. 

For these reasons, we remove the encoder and directly learn the three latent codes $\{\mathbf{Z}_i\}_{i=1}^3$ for each scene. More specifically, we initialize the $\{\mathbf{Z}_i\}_{i=1}^3$ as a tensor of zeros, and jointly optimize $\{\mathbf{Z}_i\}_{i=1}^3$ with the decoder $D$ and the entropy model $P$. At decoding time, the receiver thus receives a binary code along with the entropy model and decoder to reconstruct the sample, all three of which are counted towards the bitrate.   

\vspace{-2mm}

\paragraph{Architecture design.}  Since we must transmit the decoder $D$ along with the latent code $\{\mathbf{Z}_i\}_{i=1}^3$ to decompress the scene, it's essential for the decoder to be lightweight. \citet{yang2023computationally} established a lightweight decoder for neural image compression. We found that a two-layer transposed convolutional neural network with SELU activation \cite{klambauer2017self} is effective for our needs.

\subsection{Importance-weighted training loss} Our model is trained end-to-end (on top of the pre-trained NeRF) with a rate-distortion loss. The rate loss is defined as the log-likelihood of the entropy model $P$, and it ensures that the compressed feature planes have low relative entropy to the prior $P$. For the distortion loss, we discover that using only the NeRF rendering loss $\mathcal{L}_{render}$ is not sufficient; we also need to use an L2 feature plane reconstruction loss for good rendering quality.  

However, reconstructing the entire feature planes is not the most efficient approach for compression. Prior research \citep{li2023compressing, li2023compact, deng2023compressing} has illustrated that these feature grids possess significant redundancy and could be pruned to decrease the size of the model. Consequently, if we were to reconstruct every single grid location, it would inevitably lead to additional storage costs. 

To address this issue, we suggest computing weight maps, defined below, that we use to re-weight the feature plane reconstruction loss. With this approach, our model is guided to reconstruct only high-density grid locations while ignoring the less populated ones, ensuring a more optimized and effectively compressed representation.

For each feature plane $\mathbf{P}_i\ \in \mathbb{R}^{C_i \times W_i \times H_i}$, we define $\mathbf{W}_i \in \mathbb{R}^{W_i \times H_i}$ as the corresponding weight map, shared across all feature channels. These weight maps are constructed based on the rendering importance score $\{\mathbf{I}_i\}_{i=1}^3$ by \citet{li2023compressing, li2023compact}, defined next. 

As follows, we consider feature plane $i$ and grid location $j$. The collection of sampling points $\mathbf{x}_k \in \mathbb{R}^3$ in the vicinity of location $j$ (upon projection) shall be denoted as $\mathcal{N}_j$. Since the coordinates of $\mathbf{x}_k$ are continuous and the grid locations discrete, we distribute the "mass" of each $\mathbf{x}_k$ onto the relevant grid locations using bilinear interpolation, resulting in the interpolation weights $\omega_{kj}^i$ for sampling point $\mathbf{x}_k \in \mathcal{N}_j$. In addition, each sampling point $\mathbf{x}_k$ in Eq.~\ref{eq:rendering} has a corresponding transmittance coefficient $T_k \cdot \alpha_k$ that we interpret as its \emph{importance}. This lead to the following importance scores for each grid location $j$ in plane $i$,

\begin{equation}
    \mathbf{I}_i^{j} = \sum_{k \in \mathcal{N}_j} \omega_{kj}^i \cdot T_k \cdot \alpha_k
    \label{eq:importance_I}
\end{equation}
In sum, each importance score $\mathbf{I}_i^{j}$ is a weighted aggregate of the individual importance scores of the neighboring sampling points $\mathbf{x}_k$ over the feature grid.

Finally, we apply a log-transform to the importance maps $\{\mathbf{I}_i\}_{i=1}^3$, and then normalize them to the range of $[0, 1]$ to get the weights $\{\mathbf{W}_i\}_{i=1}^3$:
\begin{equation}
    \mathbf{W}_i = \text{normalize}(\log(\mathbf{I}_i + \epsilon)),
    \label{eq:importance_W}
\end{equation}
in which $\epsilon = 0.01$ to ensure that the log is well-defined.

\subsection{Masked entropy model}

Applying neural compression to TensoRF-VM enables us to use a wide range of different entropy models. In this section, we design a simple but effective entropy model that works well for TensoRF-VM compression, exploiting the spatial sparsity of the feature plane representation. 

Theoretically, a learned entropy model, $P$, should result in a close-to-optimal coding strategy, provided the model is flexible enough. In practice, we observed that a predominant portion of the learned latent code is zero, especially in the background. This observation might be attributed to our choice of initializing the latent codes as zero tensors and the fact that large parts of the feature planes are not used for rendering. Such sparsity is poorly captured using the standard entropy models used in neural image compression~\citep{balle2016end, balle2018variational}, leading to entropy coding inefficiencies. 

To design a better entropy model, we construct a spike-and-slab prior, oftentimes used in Bayesian statistics~\citep{mitchell1988bayesian}. To this end, we construct binary masks $\{\mathbf{M}_i\}_{i=1}^3$ into our entropy model $P$. The model $P$ compresses grid features $\mathbf{P}_i^j$ only when $\mathbf{M}_i^j = 1$, allowing selective compression of specific features and avoiding others. Those masks are learnable and can be treated as additional parameters of $P$.

In more detail, we design $P$ to be a fully factorized probability distribution as in \citet{balle2016end} and \citet{balle2018variational}. Every grid location is independent and identically distributed by binary mixture, consisting of a non-parametric distribution $p_\theta(\cdot)$ with learnable $\theta$, and a Dirac mass $\delta(\cdot)$ at zero. For each latent code $\hat{\mathbf{Z}}_i$ to be compressed, we establish a corresponding binary mask $\mathbf{M}_i$ that has the same spatial size and is shared across features channels. The conditional probability distribution $P$ (given the mask) is then factorized across spatial locations $j$ as:

\begin{equation}
\begin{aligned}
P_{\mathbf{M}_i}(\hat{\mathbf{Z}}_i) &= \prod_j p(\hat{\mathbf{Z}}_i^j | \mathbf{M}_i^j); \\
p(\hat{\mathbf{Z}}_i^j | \mathbf{M}_i^j)
 & = \begin{cases}
 \delta(\hat{\mathbf{Z}}_i^j) & \textrm {if}  \quad  \mathbf{M}_i^j = 0 \\ 
  p_\theta(\hat{\mathbf{Z}}_i^j) &  \textrm {if}\quad  \mathbf{M}_i^j = 1.
 \end{cases}
\end{aligned}
\end{equation}
We stress that we could also entropy-code the masks under a hyperprior $p({\bf M})$, but found little benefit to do so in practice. 

This construction implies that, if $\mathbf{M}_i^j = 0$, then we designate  $\hat{\mathbf{Z}}_i^j = 0$. Thus, the input to the decoder $D$ can be calculated as $\hat{\mathbf{Z}}_i \odot \mathbf{M}_i$, and the reconstructed planes are
\begin{equation}
\hat{\mathbf{P}}_i = D(\hat{\mathbf{Z}}_i \odot \mathbf{M}_i).
\label{eq:new_P}
\end{equation}

However, since the masks $\mathbf{M}_i$ are binary, they cannot be learned directly. To address this, we turn to the Gumbel-Softmax trick \citep{jang2016categorical, yang2020improving} to facilitate the learning of $\mathbf{M}_i$. For each $\mathbf{M}_i$, we define the binary probabilities denoted by $\pi_{\mathbf{M}_i}^0$ and $\pi_{\mathbf{M}_i}^1$ indicating $\mathbf{M}_i=0/1$, respectively. 
At training time, we sample $\mathbf{M}_i$ using the straight-through Gumbel-Softmax estimator \citep{bengio2013estimating, jang2016categorical}:
\begin{equation}
    \mathbf{M}_i = \argmax_{j \in \{ 0, 1 \}} (g_j + \log \pi_{\mathbf{M}_i}^j)
    \label{eq:sampling_M}
\end{equation}
in which $g_j$ are i.i.d samples drawn from $\text{Gumbel}(0, 1)$. The straight-through Gumbel-Softmax estimator allows us to calculate the gradients of $\pi_{\mathbf{M}_i}^j$. We then optimize the mask probabilities $\pi_{\mathbf{M}_i}^j$ following the rate-distortion loss:
\begin{equation}
\begin{aligned}
\mathcal{L} = & \mathcal{L}_{render}(\{\hat{\mathbf{P}}_i\}_{i=1}^3) \\
&+ \sum_{i=1}^3 \left (||(\mathbf{P}_i - \hat{\mathbf{P}}_i)\odot \mathbf{W}_i ||_2^2  - \lambda \log_2 P_{\mathbf{M}_i}(\hat{\mathbf{Z}}_i) \right),
\label{eq:final_loss}
\end{aligned}
\end{equation}
where $\hat{\mathbf{P}}_i$ is calculated with Equation \ref{eq:new_P}. In practice, we use an annealing softmax temperature $\tau$ that decays from $10$ to $0.1$ to calculate the softmax gradients. 

\section{Experiments}
\label{sec:exp}
As follows, we empirically demonstrate that our proposed approach of learning a lightweight, per-scene neural compression model without an encoder outperforms existing approaches based on vector quantization and models trained on multiple scenes in terms of rate-distortion performance.

\subsection{Experiment Setting} 
\paragraph{Datasets.} We perform our experiments on 4 datasets:
\begin{itemize}
    \item Synthetic-NeRF \citep{mildenhall2021nerf}: This dataset contains 8 scenes at resolution $800 \times 800$ rendered by Blender. Each scene contains 100 training views and 200 testing views.
    \item Synthetic-NSVF \citep{liu2020neural}: This dataset also contains 8 rendered scenes at resolution $800 \times 800$. However Synthetic-NSVF contains more complex geometry and lightning effects compared to Synthetic-NeRF.
    \item LLFF \citep{mildenhall2019local}: LLFF contains 8 real-world scenes made of forward-facing images with non empty background. We use the resolution $1008 \times 756$.
    \item Tanks and Temples \citep{knapitsch2017tanks}: We use 5 real-world scenes: \textit{Barn, Caterpillar, Family, Ignatus, Truck} from the Tanks and Temples dataset to experiment with. They have the resolution of $1920 \times 1080$.
\end{itemize}

\vspace{-1mm}

In our compression experiments, we initially train a TensoRF-VM model for every scene within the datasets listed above. We use the default TensoRF-VM 192 hyperparameters, as detailed in \citep{chen2022tensorf}. Subsequently, we apply our proposed method to compress these trained models. All experimental procedures are executed using PyTorch \citep{paszke2019pytorch} on NVIDIA RTX A6000 GPUs.

\vspace{-1mm}

\paragraph{Baselines.} We compare our compression paradigm with: The original NeRF model with MLP \citep{mildenhall2021nerf}, the uncompressed TensoRF-CP and TensoRF-VM from \citet{chen2022tensorf}, two prior compression methods for TensoRF-VM based on pruning and vector quantization named \textbf{VQ-TensoRF} from \citet{li2023compressing} and \textbf{Re:TensoRF} from \citet{deng2023compressing}.

\paragraph{Hyperparameters.}As discussed in Section \ref{sec3.1}, our decoder has two transposed convolutional layers with SELU activation \citep{klambauer2017self}. They both have a kernel size of $3$, with stride $2$ and padding $1$. Thus, each layer has an upsampling factor of $2$. Given a feature plane sized $C_i \times W_i \times H_i$, we initialize the corresponding latent code $\mathbf{Z}_i$ to have the size of $C_{Z_i} \times W_i / 4 \times H_i / 4$. 

Having a decoder with more parameters will enhance the model's decoding ability while also increase its size. In light of this trade-off, we introduce two configurations: \textbf{ECTensoRF-H} (stands for Entropy Coded TensoRF - high compression) employs latent codes with $192$ channels and a decoder with $96$ hidden channels, while \textbf{ECTensoRF-L} (low compression) has $384$ latent channels and $192$ decoder hidden channels. Regarding the hyperparameter $\lambda$, we experiment within the set $\{0.02, 0.01, 0.005, 0.001, 0.0005, 0.0002, 0.0001\}$, with higher $\lambda$ signifying a more compact model.

We train our models for $30,000$ iterations with Adam optimizer \citep{kingma2014adam}. We use an initial learning rate of $0.02$ for the latent codes and $0.001$ for the networks, and apply an exponential learning rate decay.

\begin{table*}[t]
\caption{Quantitative results comparing our method versus the baselines. PSNR is measured in dB, while the sizes are in MB. We choose the $\lambda$ to balance between the reconstruction quality and storage size.}
\label{table:quant}
\begin{center}
\resizebox{\textwidth}{!}{
\renewcommand{\arraystretch}{1.2}%
\begin{tabular}{llcccccccccccc}
\hline
 &\multicolumn{1}{c}{\multirow{2}{*}{Methods}} & \multicolumn{3}{c}{Synthetic-NeRF} & \multicolumn{3}{c}{Synthetic-NSVF} & \multicolumn{3}{c}{LLFF} & \multicolumn{3}{c}{Tanks and Temples} \\
 &\multicolumn{1}{c}{}                         & PSNR       & SSIM       & Size     & PSNR       & SSIM       & Size     & PSNR   & SSIM  & Size  & PSNR        & SSIM        & Size      \\ \hline
 \multirow{3}{*}{\rotatebox{90}{Uncom-} \rotatebox{90}{pressed}}&NeRF                                         & 31.01      & 0.947      & 5.0      & -          & -          & -        & 26.50  & 0.811  & 5.0    & 25.78       & 0.864       & 5.0       \\
 &TensoRF-CP                                   & 31.56      & 0.949      & 3.9      & 34.48      & 0.971      & 3.9      & -      & -      & -      & 27.59       & 0.897       & 3.9       \\
 &TensoRF-VM                                   & 33.09      & 0.963      & 67.6     & 36.72      & 0.982      & 71.6     & 26.70  & 0.836  & 179.8  & 28.54       & 0.921       & 72.6      \\ \hline
 \multirow{4}{*}{\rotatebox{90}{ Com-} \rotatebox{90}{pressed}}&VQ-TensoRF                                & 32.86      & 0.960      & 3.6      & 36.16      & 0.980      & 4.1      & 26.46  & 0.824  & 8.8    & 28.20       & 0.913       & 3.3       \\
 &Re:TensoRF                                & 32.81      & 0.956      & 7.9      & 36.14      & 0.978      & 8.5      & 26.55  & 0.797  & 20.2   & 28.24       & 0.907       & 6.7       \\
 & \textbf{TC-TensoRF-L (ours)}& \textbf{32.93}      & \textbf{0.961}      & \textbf{3.4}      & \textbf{36.34}      & \textbf{0.980}      & \textbf{4.0}      & \textbf{26.44}  & \textbf{0.826}  & \textbf{4.9} & \textbf{28.42}       & \textbf{0.915}       & \textbf{2.9}\\
 &TC-TensoRF-H (ours)&            32.31&            0.956&          1.6&            35.33&            0.974&          1.6&        25.72&        0.786&        1.7&             28.08&             0.907&           1.6\\ \hline
\end{tabular}
}
\end{center}
\end{table*}
\begin{table*}[h]
    \centering
    \caption{Relative improvement of our method versus VQ-TensoRF. BD-PSNR and BD-rate measure the average difference in PSNR and bitrate between the two methods.}
    \begin{tabular}{lccc}
        \hline
        & Synthetic-NeRF & Synthetic-NSVF & Tanks and Temples \\
        \hline
        BD-PSNR & 0.279 dB & 0.289 dB & 0.344 dB \\
        BD-Rate & 28.827 \% & 21.104 \% & 16.717 \% \\
        \hline
    \end{tabular}
    \label{table:bd}
\end{table*}

\subsection{Results}
We first compare our results with the baselines quantitatively. We use the PSNR and SSIM \citep{wang2004image} metrics to evaluate the reconstruction quality. The compression rate is determined by the compressed file size in MB.

\paragraph{Quantitative Results.} Table \ref{table:quant} showcases quantitative results in both rate and distortion in the high-quality/low-distortion regime, where the reconstruction quality of all compressed TensoRF models are close to other uncompressed performances for novel view synthesis. This regime is particularly relevant in NeRF compression applications, where high-quality renderings for compressed models are typically expected. 

Compared to the other two TensoRF compression baselines, VQ-TensoRF and Re:TensoRF, our variant ECTensoRF-L shows superior reconstruction performance in this regime in terms of both the PSNR and SSIM metrics while simultaneously maintaining a reduced file size across 3 datasets: Synthetic-NeRF, Synthetic-NSVF, and Tanks \& Temples. In the case of the LLFF dataset, we are slightly behind VQ-TensoRF and Re:TensoRF in PSNR. Despite this, our achieved SSIM values surpass both baselines and, remarkably, the size of our compressed files is just about half of VQ-TensoRF and a mere quarter when compared to Re:TensoRF. For a smaller number of channels, our ECTensoRF-H is able to compress the model sizes to less than $2$MB while maintaining a decent reconstruction quality. Notably, our ECTensoRF-H has a similar SSIM as Re:TensoRF on Synthetic-NeRF and Tanks\&Temples.

\vspace{-2mm}

\begin{figure*}[!t]
\begin{center}
\includegraphics[width=0.75\textwidth]{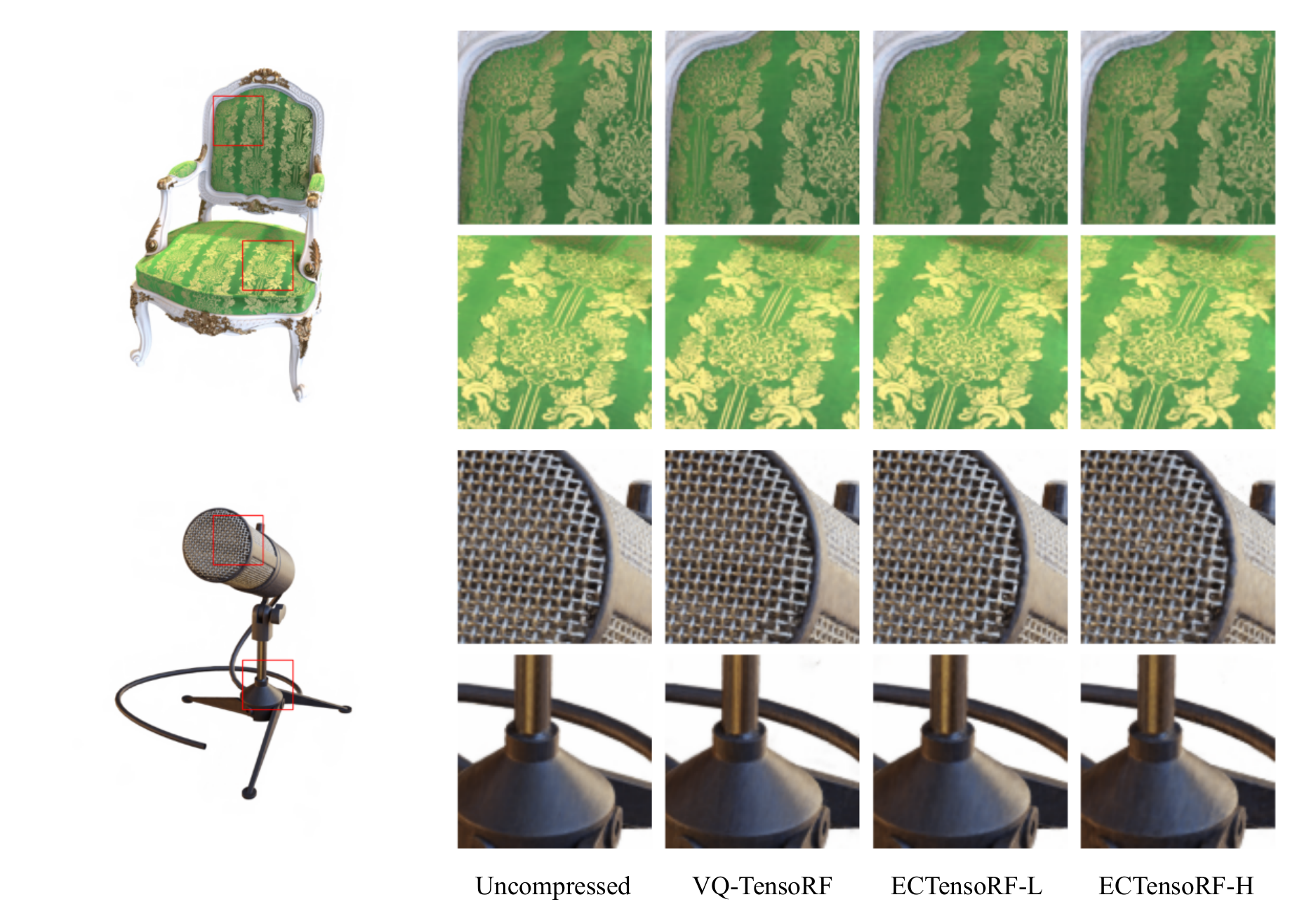}
\end{center}
\vspace{-5mm}
\caption{Qualitative results on Chair and Mic scenes from the Synthetic-NeRF dataset. From left to right: uncompressed, VQ-TensorF (average size 3.6 MB), ECTensoRF-L (3.4 MB), ECTensoRF-H (1.6 MB). Our decompressed renderings are barely distinguishable in quality from both uncompressed and VQ-compressed versions at a significantly reduced file size.}
\label{fig:qual1}
\end{figure*}

\paragraph{Qualitative Results.} We compare rendered images from the Synthetic-NeRF dataset, using VQ-TensoRF and our conpression method for both configurations: ECTensoRF-L and ECTensoRF-H in Figure \ref{fig:qual1}. Visually, there is minimal disparity between the uncompressed and compressed TensoRF models. We further show more qualitative results for the other datasets in the Appendix.

\paragraph{Rate-distortion performance.} The rate-distortion curve is widely used in neural compression to compare the compression performance across different compression level. Here we analyze the rate-distortion curve of our ECTensoRF-L with various $\lambda$ values versus VQ-TensoRF with various codebook size. For the VQ-TensoRF evaluations, we employed the officially released code and utilized the same pre-trained TensoRF models for consistency. 

\begin{figure}[!t]
\begin{center}
\vspace{-7mm}
\includegraphics[width=0.47\textwidth]{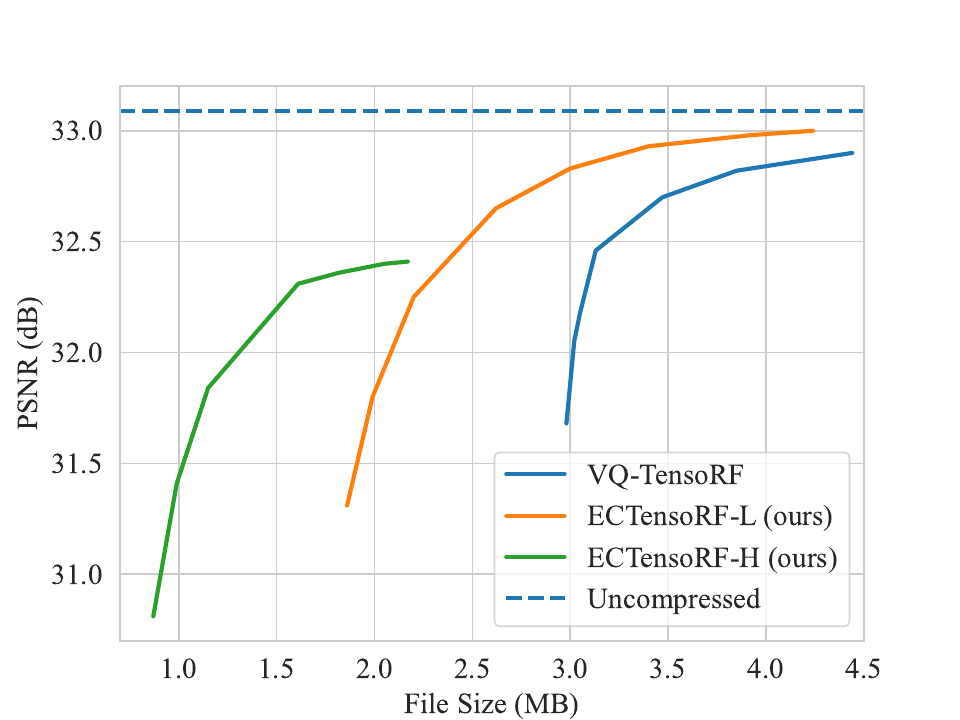}
\includegraphics[width=0.47\textwidth]{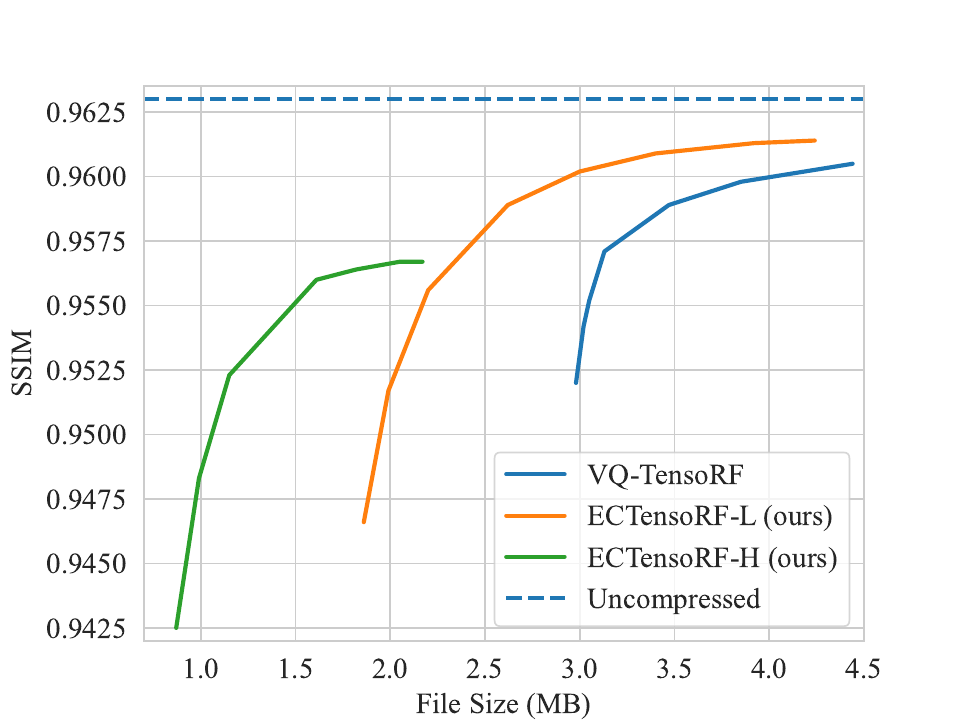}
\vspace{-1.9mm}
\caption{Comparison of rate-distortion curves between our proposed methods and the baseline VQ-TensoRF on the Synthetic-NeRF dataset. The upper figure illustrates PSNR against file size, and the lower figure showcases SSIM in relation to file size.}
\vspace{-12mm}
\label{fig:rd_psnr_nerf}
\end{center}
\end{figure}

Figure \ref{fig:rd_psnr_nerf} shows that our ECTensoRF-L outpaces VQ-TensoRF across various levels of compression in Synthetic-NeRF dataset with both PSNR and SSIM metrics. Rate-distortion curves for other datasets can be found in the Appendix \ref{sec:appendix2}.

Moreover, Table \ref{table:bd} shows the relative improvement of our method over VQ-TensoRF using Bjontegaard Delta (BD) BD-PSNR and BD-rate metrics \citep{bjontegaard2001calculation}, highlighting that our model achieves better PSNR and bit-rate across various compression levels.

\vspace{-1mm}

\paragraph{Training and rendering time.} Training an uncompressed TensoRF model for a scene from the Synthetic-NeRF dataset takes around 15 minutes on an NVIDIA A6000 GPU. Running on top of that, our compression method takes an additional 40 minutes. Our framework is slower than the baseline VQ-TensoRF, which runs in 7 minutes on the same hardware. Regarding rendering, our approach adds a negligible overhead of roughly 2 seconds for the decompression of parameters. Once decompressed, the rendering procedure is the same as TensoRF.

\vspace{-1mm}

\paragraph{Compression details.} The average storage size breakdown of our model on the Synthetic-NeRF dataset (with the configuration from Table \ref{table:quant}) is provided in the Table \ref{table:storage_size}. For the feature planes, we compresse them with the learned entropy model. All the other components (the renderer MLP, decoder, density/appearance vectors, learned masks, entropy bottleneck parameters and model config) are packed into a single file and compressed with LZ77 \citep{ziv1977universal}.

\begin{table}[t]
\centering
\caption{Storage size breakdown.}
\begin{tabular}{lc}
\hline
Component & Size (MB) \\
\hline
Feature planes & 1.657 \\
Decoder & 1.380 \\
Other components & 0.366 \\
\hline
Total & 3.403 \\
\hline
\end{tabular}
\label{table:storage_size}
\end{table}

\section{Ablation Studies}
We conduct experiments to verify our design choices. We test on the Synthetic-NeRF datasets, with our ECTensoRF-L architecture. 
\subsection{Advantages of per-scene optimization}
As outlined in Section \ref{sec3.1}, our compression methodology is optimized on a per-scene basis. However, this raises the question: how is the performance of traditional nonlinear transform coding on TensoRF compression, even on a small-scale dataset? To address this, we conduct a comparative analysis. We compress the TensoRF model by first training a compression network using an encoder-decoder architecture, similar to traditional nonlinear transform coding. Specifically, we train the compression network on seven scenes from the Synthetic-NeRF dataset and tested the trained model on the remaining scene (Lego). We compare this approach with our per-scene optimized model.
\begin{figure}[!t]
\begin{center}
\includegraphics[width=0.43\textwidth]{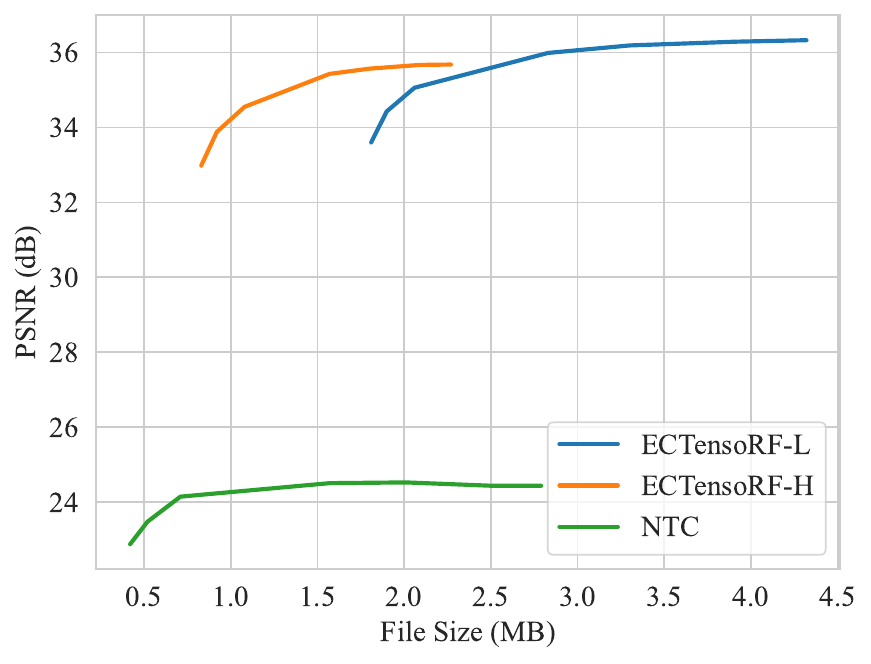}
\caption{Rate-distortion comparison between traditional nonlinear transform coding (green), trained across 7 scenes, and our per-scene compression methods (orange, blue).}
\label{fig:rd_ab_ntc}
\end{center}
\end{figure}

Figure \ref{fig:rd_ab_ntc} shows the results of this experiment. Compared to per-scene training, pre-trained NTC suffers from inferior reconstruction quality. More specifically, the maximum PSNR that the pre-trained NTC can achieve is only 24.52 dB, which is 12.03 dB lower than the uncompressed PSNR value (36.55 dB) and also much lower than the PSNR values of per-scene trained models. However, we note that the major advantage of pre-trained NTC is a much faster compression time. Using a pre-trained NTC model also avoids the need to transmit the entropy model and the decoder, as we assume that the receiver always has access to them, which is similar to the image compression setting.

\subsection{Ablation on other design choices}
\begin{figure}[!t]
\begin{center}
\includegraphics[width=0.45\textwidth]{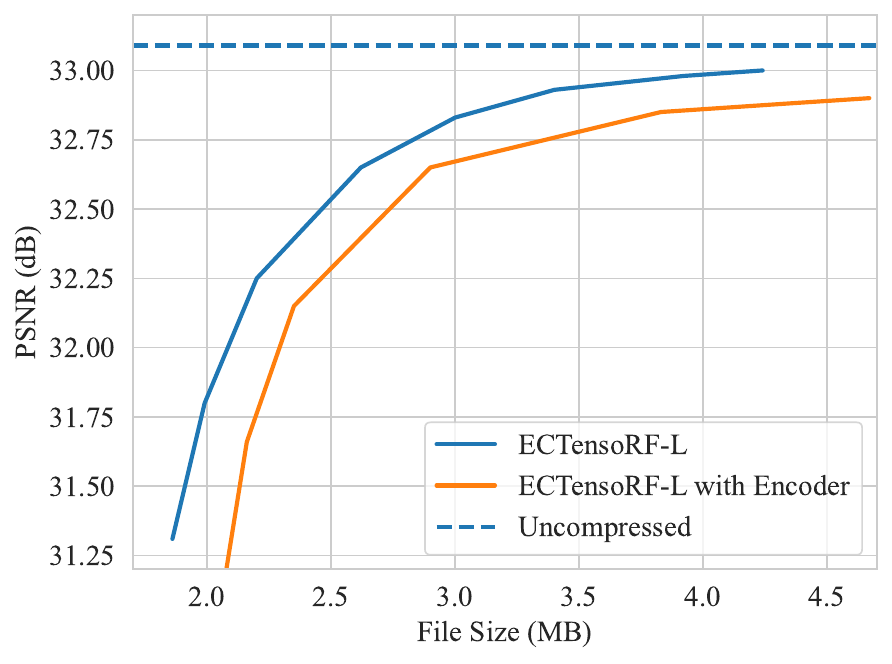}
\includegraphics[width=0.45\textwidth]{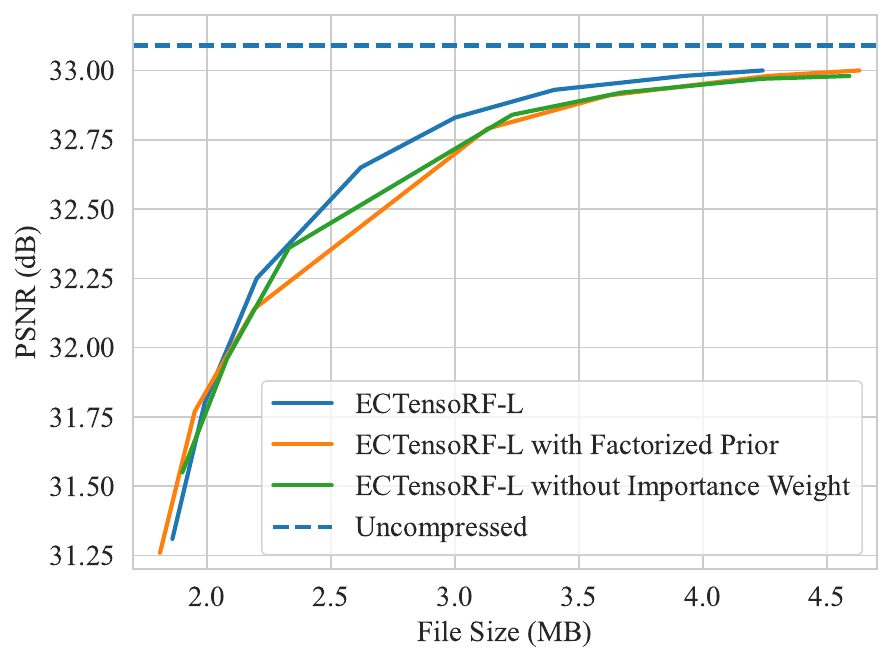}
\caption{Ablation studies. 
Top: rate-distortion comparison of our approach against a version with encoder, trained on a single scene. 
Bottom: comparisons between model versions with factorized prior and without importance weight.
}
\label{fig:rd_ablation1}
\end{center}
\end{figure}

\paragraph{Using the encoder} We first show the sub-optimal performance of ECTensoRF-L compression with an encoder. As discussed in Section 3.1, using an encoder leads to an irreducible amortization gap in optimization, and the resulting compression performance is worse, as shown in Figure \ref{fig:rd_ablation1}.

\paragraph{Training without Importance-Weighted Loss.} We examine the rate-distortion curves of ECTensoRF-L, trained both with and without importance weight, as depicted in Figure \ref{fig:rd_ablation1}. At an identical PSNR of $32.98$ dB, employing importance weight in training our model helps reduce the file size from $4.59$ MB to $3.92$ MB.

\paragraph{The Effect of the Masked Entropy Model.} To demonstrate the efficacy of our masked entropy model, we undertook a comparative analysis between the compression performance of ECTensoRF-L using the conventional factorized prior \citep{balle2016end, balle2018variational} and our masked model. The results related to rate distortion curves can be found in the bottom plot of Figure \ref{fig:rd_ablation1}.

It's noteworthy that, due to the additional overhead introduced by sending the masks, our results lag slightly behind the factorized prior in a low-rate setting. Yet, in medium to high-rate regimes, our prior emerges superior compared to the traditional factorized prior. To illustrate, for a PSNR value of $32.98$ dB, the compressed file with the factorized prior occupies $4.26$ MB. In contrast, our method employing the proposed masked entropy model results in a reduced file size of $3.92$ MB.

\begin{figure}[!t]
\begin{center}
\includegraphics[width=0.48\textwidth]{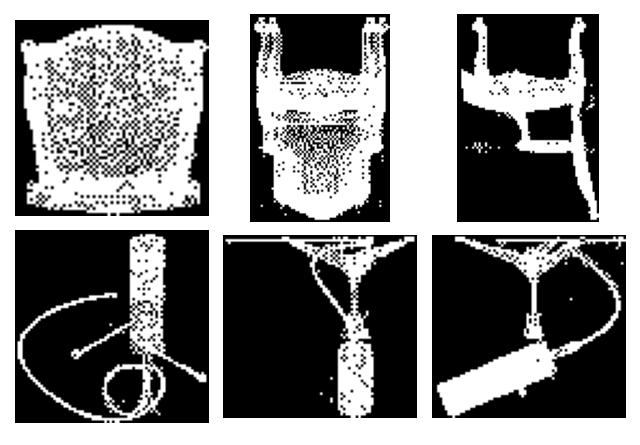}
\caption{Ablation studies. We show the sparsity masks of our entropy model learned on the Chair and Mic scene.}
\label{fig:rd_ablation2}
\end{center}
\end{figure}

To further understand the behavior of our masked entropy model, we visualize the masks learned for the Chair and Mic scene from Synthetic-NeRF dataset in Figure \ref{fig:rd_ablation2}.  We can observe that the masks resemble the rendering objects when viewed from different angles, and they inherently ignore the background. This behavior is similar to the pruning strategies employed in prior grid-based NeRF compression works \citep{li2023compressing, li2023compact, deng2023compressing}.

\paragraph{More experimental results.} We further conduct experiments on different latents initialization and end-to-end training in the Appendix \ref{sec:appendix2}. 

\section{Related Works and Discussion}
\paragraph{Grid-based NeRF compression.} Since storage cost is a significant challenge of grid-based NeRF, several methods were proposed to solve this problem. \citet{li2023compressing} introduces a three-stage approach, integrating voxel pruning and vector quantization \citep{gray1984vector} through a learnable codebook. Similarly, Re:NeRF \citep{deng2023compressing} employs voxel pruning, but adopts a strategy of sequentially removing and reintegrating parameters to prevent a significant drop in performance. Meanwhile, \citet{takikawa2022variable} adopts the codebook idea from Instant-NGP \citep{muller2022instant}, but substitutes hash-based encoding with a learned mapping that associates grid positions to corresponding codebook indices. However this approach requires considerable training memory. \citet{li2023compact} applies downsampling to the voxels and employs a network to enhance render quality. Our method shares some similarity to \citet{li2023compact}, but we learn the downsampled latent codes with a novel entropy model to effectively compress them. Additionally, while our masked factorized prior also resembles the pruning mechanism used in previous works, our method differentiates itself by adaptively learning the masks instead of relying on fixed thresholds.

\vspace{-2mm}

\paragraph{Neural compression for NeRF.} Applying neural compression to NeRF is a relatively young field. \citet{bird20213d} learns an entropy model to compress the MLP-based NeRF \cite{mildenhall2021nerf} network weights, based on the prior model compression work of \citet{oktay2019scalable}. In contrast, our work focuses on compressing the feature grids of grid-based NeRF. We additionally improve the conventional compression procedure and propose a novel entropy model. Concurrent to our work, \citet{li2024nerfcodec} also applies neural compression to TensoRF by leveraging a pretrained image comression network.

\vspace{-2mm}

\paragraph{Discussion.} Throughout this paper,  our emphasis has been on applying neural compression techniques specifically to TensoRF. Nonetheless, our method has the potential to be applied to other grid-based NeRF methods beyond just TensoRF, such as Triplanes \citep{chan2022efficient, fridovich2023k}, Factor Fields \citep{chen2023factor} or DVGO \citep{sun2022direct}. Taking DVGO as an example, we can learn a 4D latent code and have an entropy model to model its probability density. Then a decoder may decode this 4D latent code to render the scene. 

\section{Conclusion}
In this study, we present a novel approach to applying neural compression to the TensoRF model, a prominent grid-based NeRF method. Our approach adapts traditional neural compression techniques, commonly used in image and video compression, to NeRF models. We develop an efficient per-scene optimization scheme and propose various designs, such as importance-weighted feature reconstruction and a masked entropy model. Our experiments demonstrate that we can significantly reduce storage requirements of a NeRF model with only a minimal compromise in rendering quality, and outperform previous NeRF compression baselines. More importantly, our compression method only adds minimal overhead to the rendering process.

\vspace{-2mm}

\paragraph{Limitation and future work.} 
One limitation of our neural compression approach is the longer training time compared to the baseline VQ-TensoRF, as mentioned in Section \ref{sec:exp}. Additionally, the final compressed model still includes the cost of transmitting the decoder. Future work could focus on reducing compression time, learning a network compression model \citep{oktay2019scalable, girish2022lilnetx} to compress the decoder network, and applying our method to other NeRF architectures.

\section*{Impact Statement}
Neural compression is a collection of methods that advance data compression with end-to-end learning approaches. Biased training data may influence how models reconstruct data and may lead to misrepresentations, e.g., of individuals, especially at low bitrates.

\section*{Acknowledgements}
The authors acknowledge support from the National Science Foundation (NSF) under an NSF CAREER Award (2047418), award numbers 2003237 and 2007719, by the Department of Energy under grant DE-SC0022331, the IARPA WRIVA program, and by gifts from Qualcomm and Disney. We also thank Justus Will for his meticulous proofreading and valuable suggestions for this paper.

\nocite{langley00}

\bibliography{references}

\begin{thebibliography}{52}
\providecommand{\natexlab}[1]{#1}
\providecommand{\url}[1]{\texttt{#1}}
\expandafter\ifx\csname urlstyle\endcsname\relax
  \providecommand{\doi}[1]{doi: #1}\else
  \providecommand{\doi}{doi: \begingroup \urlstyle{rm}\Url}\fi

\bibitem[ziv(1977)]{ziv1977universal}
A universal algorithm for sequential data compression.
\newblock \emph{IEEE Transactions on information theory}, 23\penalty0 (3):\penalty0 337--343, 1977.

\bibitem[Ball{\'e} et~al.(2016)Ball{\'e}, Laparra, and Simoncelli]{balle2016end}
Ball{\'e}, J., Laparra, V., and Simoncelli, E.~P.
\newblock End-to-end optimization of nonlinear transform codes for perceptual quality.
\newblock In \emph{2016 Picture Coding Symposium (PCS)}, pp.\  1--5. IEEE, 2016.

\bibitem[Ball{\'e} et~al.(2018)Ball{\'e}, Minnen, Singh, Hwang, and Johnston]{balle2018variational}
Ball{\'e}, J., Minnen, D., Singh, S., Hwang, S.~J., and Johnston, N.
\newblock Variational image compression with a scale hyperprior.
\newblock In \emph{International Conference on Learning Representations}, 2018.

\bibitem[Ball{\'e} et~al.(2020)Ball{\'e}, Chou, Minnen, Singh, Johnston, Agustsson, Hwang, and Toderici]{balle2020nonlinear}
Ball{\'e}, J., Chou, P.~A., Minnen, D., Singh, S., Johnston, N., Agustsson, E., Hwang, S.~J., and Toderici, G.
\newblock Nonlinear transform coding.
\newblock \emph{IEEE Journal of Selected Topics in Signal Processing}, 15\penalty0 (2):\penalty0 339--353, 2020.

\bibitem[Barron et~al.(2021)Barron, Mildenhall, Tancik, Hedman, Martin-Brualla, and Srinivasan]{barron2021mip}
Barron, J.~T., Mildenhall, B., Tancik, M., Hedman, P., Martin-Brualla, R., and Srinivasan, P.~P.
\newblock Mip-nerf: A multiscale representation for anti-aliasing neural radiance fields.
\newblock In \emph{Proceedings of the IEEE/CVF International Conference on Computer Vision}, pp.\  5855--5864, 2021.

\bibitem[Bengio et~al.(2013)Bengio, L{\'e}onard, and Courville]{bengio2013estimating}
Bengio, Y., L{\'e}onard, N., and Courville, A.
\newblock Estimating or propagating gradients through stochastic neurons for conditional computation.
\newblock \emph{arXiv preprint arXiv:1308.3432}, 2013.

\bibitem[Bird et~al.(2021)Bird, Ball{\'e}, Singh, and Chou]{bird20213d}
Bird, T., Ball{\'e}, J., Singh, S., and Chou, P.~A.
\newblock 3d scene compression through entropy penalized neural representation functions.
\newblock In \emph{2021 Picture Coding Symposium (PCS)}, pp.\  1--5. IEEE, 2021.

\bibitem[Bjontegaard(2001)]{bjontegaard2001calculation}
Bjontegaard, G.
\newblock Calculation of average psnr differences between rd-curves.
\newblock \emph{ITU SG16 Doc. VCEG-M33}, 2001.

\bibitem[Campos et~al.(2019)Campos, Meierhans, Djelouah, and Schroers]{campos2019content}
Campos, J., Meierhans, S., Djelouah, A., and Schroers, C.
\newblock Content adaptive optimization for neural image compression.
\newblock In \emph{Proceedings of the IEEE/CVF Conference on Computer Vision and Pattern Recognition Workshops}, pp.\  0--0, 2019.

\bibitem[Chan et~al.(2022)Chan, Lin, Chan, Nagano, Pan, De~Mello, Gallo, Guibas, Tremblay, Khamis, et~al.]{chan2022efficient}
Chan, E.~R., Lin, C.~Z., Chan, M.~A., Nagano, K., Pan, B., De~Mello, S., Gallo, O., Guibas, L.~J., Tremblay, J., Khamis, S., et~al.
\newblock Efficient geometry-aware 3d generative adversarial networks.
\newblock In \emph{Proceedings of the IEEE/CVF Conference on Computer Vision and Pattern Recognition}, pp.\  16123--16133, 2022.

\bibitem[Chen et~al.(2022)Chen, Xu, Geiger, Yu, and Su]{chen2022tensorf}
Chen, A., Xu, Z., Geiger, A., Yu, J., and Su, H.
\newblock Tensorf: Tensorial radiance fields.
\newblock In \emph{European Conference on Computer Vision}, pp.\  333--350. Springer, 2022.

\bibitem[Chen et~al.(2023)Chen, Xu, Wei, Tang, Su, and Geiger]{chen2023factor}
Chen, A., Xu, Z., Wei, X., Tang, S., Su, H., and Geiger, A.
\newblock Factor fields: A unified framework for neural fields and beyond.
\newblock \emph{arXiv preprint arXiv:2302.01226}, 2023.

\bibitem[Cheng et~al.(2020)Cheng, Sun, Takeuchi, and Katto]{cheng2020image}
Cheng, Z., Sun, H., Takeuchi, M., and Katto, J.
\newblock Learned image compression with discretized gaussian mixture likelihoods and attention modules.
\newblock In \emph{Proceedings of the IEEE Conference on Computer Vision and Pattern Recognition (CVPR)}, 2020.

\bibitem[Cremer et~al.(2018)Cremer, Li, and Duvenaud]{cremer2018inference}
Cremer, C., Li, X., and Duvenaud, D.
\newblock Inference suboptimality in variational autoencoders.
\newblock In \emph{International Conference on Machine Learning}, pp.\  1078--1086. PMLR, 2018.

\bibitem[Deitke et~al.(2023)Deitke, Schwenk, Salvador, Weihs, Michel, VanderBilt, Schmidt, Ehsani, Kembhavi, and Farhadi]{deitke2023objaverse}
Deitke, M., Schwenk, D., Salvador, J., Weihs, L., Michel, O., VanderBilt, E., Schmidt, L., Ehsani, K., Kembhavi, A., and Farhadi, A.
\newblock Objaverse: A universe of annotated 3d objects.
\newblock In \emph{Proceedings of the IEEE/CVF Conference on Computer Vision and Pattern Recognition}, pp.\  13142--13153, 2023.

\bibitem[Deitke et~al.(2024)Deitke, Liu, Wallingford, Ngo, Michel, Kusupati, Fan, Laforte, Voleti, Gadre, et~al.]{deitke2024objaverse}
Deitke, M., Liu, R., Wallingford, M., Ngo, H., Michel, O., Kusupati, A., Fan, A., Laforte, C., Voleti, V., Gadre, S.~Y., et~al.
\newblock Objaverse-xl: A universe of 10m+ 3d objects.
\newblock \emph{Advances in Neural Information Processing Systems}, 36, 2024.

\bibitem[Deng \& Tartaglione(2023)Deng and Tartaglione]{deng2023compressing}
Deng, C.~L. and Tartaglione, E.
\newblock Compressing explicit voxel grid representations: fast nerfs become also small.
\newblock In \emph{Proceedings of the IEEE/CVF Winter Conference on Applications of Computer Vision}, pp.\  1236--1245, 2023.

\bibitem[Fridovich-Keil et~al.(2022)Fridovich-Keil, Yu, Tancik, Chen, Recht, and Kanazawa]{fridovich2022plenoxels}
Fridovich-Keil, S., Yu, A., Tancik, M., Chen, Q., Recht, B., and Kanazawa, A.
\newblock Plenoxels: Radiance fields without neural networks.
\newblock In \emph{Proceedings of the IEEE/CVF Conference on Computer Vision and Pattern Recognition}, pp.\  5501--5510, 2022.

\bibitem[Fridovich-Keil et~al.(2023)Fridovich-Keil, Meanti, Warburg, Recht, and Kanazawa]{fridovich2023k}
Fridovich-Keil, S., Meanti, G., Warburg, F.~R., Recht, B., and Kanazawa, A.
\newblock K-planes: Explicit radiance fields in space, time, and appearance.
\newblock In \emph{Proceedings of the IEEE/CVF Conference on Computer Vision and Pattern Recognition}, pp.\  12479--12488, 2023.

\bibitem[Gershman \& Goodman(2014)Gershman and Goodman]{gershman2014amortized}
Gershman, S. and Goodman, N.
\newblock Amortized inference in probabilistic reasoning.
\newblock In \emph{Proceedings of the annual meeting of the cognitive science society}, volume~36, 2014.

\bibitem[Girish et~al.(2022)Girish, Gupta, Singh, and Shrivastava]{girish2022lilnetx}
Girish, S., Gupta, K., Singh, S., and Shrivastava, A.
\newblock Lilnetx: Lightweight networks with extreme model compression and structured sparsification.
\newblock In \emph{The Eleventh International Conference on Learning Representations}, 2022.

\bibitem[Gray(1984)]{gray1984vector}
Gray, R.
\newblock Vector quantization.
\newblock \emph{IEEE Assp Magazine}, 1\penalty0 (2):\penalty0 4--29, 1984.

\bibitem[Jang et~al.(2016)Jang, Gu, and Poole]{jang2016categorical}
Jang, E., Gu, S., and Poole, B.
\newblock Categorical reparameterization with gumbel-softmax.
\newblock In \emph{International Conference on Learning Representations}, 2016.

\bibitem[Kingma \& Ba(2015)Kingma and Ba]{kingma2014adam}
Kingma, D.~P. and Ba, J.
\newblock Adam: A method for stochastic optimization.
\newblock In \emph{International Conference on Learning Representations}, 2015.

\bibitem[Kingma \& Welling(2014)Kingma and Welling]{kingma2013auto}
Kingma, D.~P. and Welling, M.
\newblock Auto-encoding variational bayes.
\newblock In \emph{International Conference on Learning Representations}, 2014.

\bibitem[Klambauer et~al.(2017)Klambauer, Unterthiner, Mayr, and Hochreiter]{klambauer2017self}
Klambauer, G., Unterthiner, T., Mayr, A., and Hochreiter, S.
\newblock Self-normalizing neural networks.
\newblock \emph{Advances in neural information processing systems}, 30, 2017.

\bibitem[Knapitsch et~al.(2017)Knapitsch, Park, Zhou, and Koltun]{knapitsch2017tanks}
Knapitsch, A., Park, J., Zhou, Q.-Y., and Koltun, V.
\newblock Tanks and temples: Benchmarking large-scale scene reconstruction.
\newblock \emph{ACM Transactions on Graphics (ToG)}, 36\penalty0 (4):\penalty0 1--13, 2017.

\bibitem[Kolda \& Bader(2009)Kolda and Bader]{kolda2009tensor}
Kolda, T.~G. and Bader, B.~W.
\newblock Tensor decompositions and applications.
\newblock \emph{SIAM review}, 51\penalty0 (3):\penalty0 455--500, 2009.

\bibitem[Li et~al.(2023{\natexlab{a}})Li, Shen, Wang, Shen, and Bo]{li2023compressing}
Li, L., Shen, Z., Wang, Z., Shen, L., and Bo, L.
\newblock Compressing volumetric radiance fields to 1 mb.
\newblock In \emph{Proceedings of the IEEE/CVF Conference on Computer Vision and Pattern Recognition}, pp.\  4222--4231, 2023{\natexlab{a}}.

\bibitem[Li et~al.(2023{\natexlab{b}})Li, Wang, Shen, Shen, and Tan]{li2023compact}
Li, L., Wang, Z., Shen, Z., Shen, L., and Tan, P.
\newblock Compact real-time radiance fields with neural codebook.
\newblock In \emph{2023 IEEE International Conference on Multimedia and Expo (ICME)}, pp.\  2189--2194. IEEE, 2023{\natexlab{b}}.

\bibitem[Li et~al.(2024)Li, Li, Liao, and Yu]{li2024nerfcodec}
Li, S., Li, H., Liao, Y., and Yu, L.
\newblock Nerfcodec: Neural feature compression meets neural radiance fields for memory-efficient scene representation.
\newblock \emph{arXiv preprint arXiv:2404.02185}, 2024.

\bibitem[Liu et~al.(2020)Liu, Gu, Zaw~Lin, Chua, and Theobalt]{liu2020neural}
Liu, L., Gu, J., Zaw~Lin, K., Chua, T.-S., and Theobalt, C.
\newblock Neural sparse voxel fields.
\newblock \emph{Advances in Neural Information Processing Systems}, 33:\penalty0 15651--15663, 2020.

\bibitem[Marino et~al.(2018)Marino, Yue, and Mandt]{marino2018iterative}
Marino, J., Yue, Y., and Mandt, S.
\newblock Iterative amortized inference.
\newblock In \emph{International Conference on Machine Learning}, pp.\  3403--3412. PMLR, 2018.

\bibitem[Matsubara et~al.(2022)Matsubara, Yang, Levorato, and Mandt]{matsubara2022supervised}
Matsubara, Y., Yang, R., Levorato, M., and Mandt, S.
\newblock Supervised compression for resource-constrained edge computing systems.
\newblock In \emph{Proceedings of the IEEE/CVF Winter Conference on Applications of Computer Vision}, pp.\  2685--2695, 2022.

\bibitem[Mildenhall et~al.(2019)Mildenhall, Srinivasan, Ortiz-Cayon, Kalantari, Ramamoorthi, Ng, and Kar]{mildenhall2019local}
Mildenhall, B., Srinivasan, P.~P., Ortiz-Cayon, R., Kalantari, N.~K., Ramamoorthi, R., Ng, R., and Kar, A.
\newblock Local light field fusion: Practical view synthesis with prescriptive sampling guidelines.
\newblock \emph{ACM Transactions on Graphics (TOG)}, 38\penalty0 (4):\penalty0 1--14, 2019.

\bibitem[Mildenhall et~al.(2021)Mildenhall, Srinivasan, Tancik, Barron, Ramamoorthi, and Ng]{mildenhall2021nerf}
Mildenhall, B., Srinivasan, P.~P., Tancik, M., Barron, J.~T., Ramamoorthi, R., and Ng, R.
\newblock Nerf: Representing scenes as neural radiance fields for view synthesis.
\newblock \emph{Communications of the ACM}, 65\penalty0 (1):\penalty0 99--106, 2021.

\bibitem[Minnen et~al.(2018)Minnen, Ball{\'e}, and Toderici]{minnen2018joint}
Minnen, D., Ball{\'e}, J., and Toderici, G.~D.
\newblock Joint autoregressive and hierarchical priors for learned image compression.
\newblock \emph{Advances in neural information processing systems}, 31, 2018.

\bibitem[Mitchell \& Beauchamp(1988)Mitchell and Beauchamp]{mitchell1988bayesian}
Mitchell, T.~J. and Beauchamp, J.~J.
\newblock Bayesian variable selection in linear regression.
\newblock \emph{Journal of the american statistical association}, 83\penalty0 (404):\penalty0 1023--1032, 1988.

\bibitem[M{\"u}ller et~al.(2022)M{\"u}ller, Evans, Schied, and Keller]{muller2022instant}
M{\"u}ller, T., Evans, A., Schied, C., and Keller, A.
\newblock Instant neural graphics primitives with a multiresolution hash encoding.
\newblock \emph{ACM Transactions on Graphics (ToG)}, 41\penalty0 (4):\penalty0 1--15, 2022.

\bibitem[Oktay et~al.(2019)Oktay, Ball{\'e}, Singh, and Shrivastava]{oktay2019scalable}
Oktay, D., Ball{\'e}, J., Singh, S., and Shrivastava, A.
\newblock Scalable model compression by entropy penalized reparameterization.
\newblock In \emph{International Conference on Learning Representations}, 2019.

\bibitem[Paszke et~al.(2019)Paszke, Gross, Massa, Lerer, Bradbury, Chanan, Killeen, Lin, Gimelshein, Antiga, et~al.]{paszke2019pytorch}
Paszke, A., Gross, S., Massa, F., Lerer, A., Bradbury, J., Chanan, G., Killeen, T., Lin, Z., Gimelshein, N., Antiga, L., et~al.
\newblock Pytorch: An imperative style, high-performance deep learning library.
\newblock \emph{Advances in neural information processing systems}, 32, 2019.

\bibitem[Sun et~al.(2022)Sun, Sun, and Chen]{sun2022direct}
Sun, C., Sun, M., and Chen, H.-T.
\newblock Direct voxel grid optimization: Super-fast convergence for radiance fields reconstruction.
\newblock In \emph{Proceedings of the IEEE/CVF Conference on Computer Vision and Pattern Recognition}, pp.\  5459--5469, 2022.

\bibitem[Takikawa et~al.(2022)Takikawa, Evans, Tremblay, M{\"u}ller, McGuire, Jacobson, and Fidler]{takikawa2022variable}
Takikawa, T., Evans, A., Tremblay, J., M{\"u}ller, T., McGuire, M., Jacobson, A., and Fidler, S.
\newblock Variable bitrate neural fields.
\newblock In \emph{ACM SIGGRAPH 2022 Conference Proceedings}, pp.\  1--9, 2022.

\bibitem[Wallace(1991)]{wallace1991jpeg}
Wallace, G.~K.
\newblock The jpeg still picture compression standard.
\newblock \emph{Communications of the ACM}, 34\penalty0 (4):\penalty0 30--44, 1991.

\bibitem[Wang et~al.(2004)Wang, Bovik, Sheikh, and Simoncelli]{wang2004image}
Wang, Z., Bovik, A.~C., Sheikh, H.~R., and Simoncelli, E.~P.
\newblock Image quality assessment: from error visibility to structural similarity.
\newblock \emph{IEEE transactions on image processing}, 13\penalty0 (4):\penalty0 600--612, 2004.

\bibitem[Yang \& Mandt(2023{\natexlab{a}})Yang and Mandt]{yang2023lossy}
Yang, R. and Mandt, S.
\newblock Lossy image compression with conditional diffusion models.
\newblock \emph{Advances in Neural Information Processing Systems}, 36, 2023{\natexlab{a}}.

\bibitem[Yang et~al.(2023{\natexlab{a}})Yang, Yang, Marino, and Mandt]{yang2023insights}
Yang, R., Yang, Y., Marino, J., and Mandt, S.
\newblock Insights from generative modeling for neural video compression.
\newblock \emph{IEEE Transactions on Pattern Analysis and Machine Intelligence}, 2023{\natexlab{a}}.

\bibitem[Yang \& Mandt(2023{\natexlab{b}})Yang and Mandt]{yang2023computationally}
Yang, Y. and Mandt, S.
\newblock Computationally-efficient neural image compression with shallow decoders.
\newblock In \emph{Proceedings of the IEEE/CVF International Conference on Computer Vision}, pp.\  530--540, 2023{\natexlab{b}}.

\bibitem[Yang et~al.(2020)Yang, Bamler, and Mandt]{yang2020improving}
Yang, Y., Bamler, R., and Mandt, S.
\newblock Improving inference for neural image compression.
\newblock \emph{Advances in Neural Information Processing Systems}, 33:\penalty0 573--584, 2020.

\bibitem[Yang et~al.(2023{\natexlab{b}})Yang, Mandt, Theis, et~al.]{yang2023introduction}
Yang, Y., Mandt, S., Theis, L., et~al.
\newblock An introduction to neural data compression.
\newblock \emph{Foundations and Trends{\textregistered} in Computer Graphics and Vision}, 15\penalty0 (2):\penalty0 113--200, 2023{\natexlab{b}}.

\bibitem[Yu et~al.(2021)Yu, Li, Tancik, Li, Ng, and Kanazawa]{yu2021plenoctrees}
Yu, A., Li, R., Tancik, M., Li, H., Ng, R., and Kanazawa, A.
\newblock Plenoctrees for real-time rendering of neural radiance fields.
\newblock In \emph{Proceedings of the IEEE/CVF International Conference on Computer Vision}, pp.\  5752--5761, 2021.

\bibitem[Zhang et~al.(2020)Zhang, Riegler, Snavely, and Koltun]{zhang2020nerf++}
Zhang, K., Riegler, G., Snavely, N., and Koltun, V.
\newblock Nerf++: Analyzing and improving neural radiance fields.
\newblock \emph{arXiv preprint arXiv:2010.07492}, 2020.

\end{thebibliography}
\bibliographystyle{icml2024}

\newpage
\appendix
\onecolumn
\section{Appendix}
\subsection{Algorithm}

\begin{algorithm}[H]
    \caption{TensoRF-VM compression}
    \begin{algorithmic}

    \SetKwInOut{Input}{Input}
    \SetKwInOut{Output}{Output}
    \SetKwFunction{to}{to}
    \STATE {\bfseries Input:} Pretrained TensoRF-VM model
    \STATE {\bfseries Output:} Compressed TensoRF-VM model
    \STATE Calculate $\{\mathbf{W}_i\}_{i=1}^3$ using Eq \ref{eq:importance_I} and \ref{eq:importance_W}
    \STATE Initialize $\{\mathbf{Z}_i\}_{i=1}^3$ as $0$-tensors
    \STATE Initialize decoder $D$ and entropy model $P$\ with masks parameters $\{\pi_{\mathbf{M}_i}^0\}_{i=1}^3$ and $\{\pi_{\mathbf{M}_i}^1\}_{i=1}^3$
    \WHILE{\text{not converged}}
        \STATE Sample $\{\mathbf{M}_i\}_{i=1}^3$ using Gumbel-Softmax as in Eq \ref{eq:sampling_M}
        \STATE Reconstruct $\{\hat{\mathbf{P}}_i\}_{i=1}^3$ by Eq \ref{eq:new_P}
        \STATE Render the scene with $\{\hat{\mathbf{P}}_i\}_{i=1}^3$
        \STATE Calculate the loss in Eq \ref{eq:final_loss}\ and update the model
    \ENDWHILE
    \end{algorithmic}
\end{algorithm}

\subsection{More experimental results}
\label{sec:appendix2}
\subsubsection{Rate-distortion comparison on other datasets}
\begin{figure*}[h]
\begin{center}
\includegraphics[width=0.48\textwidth]{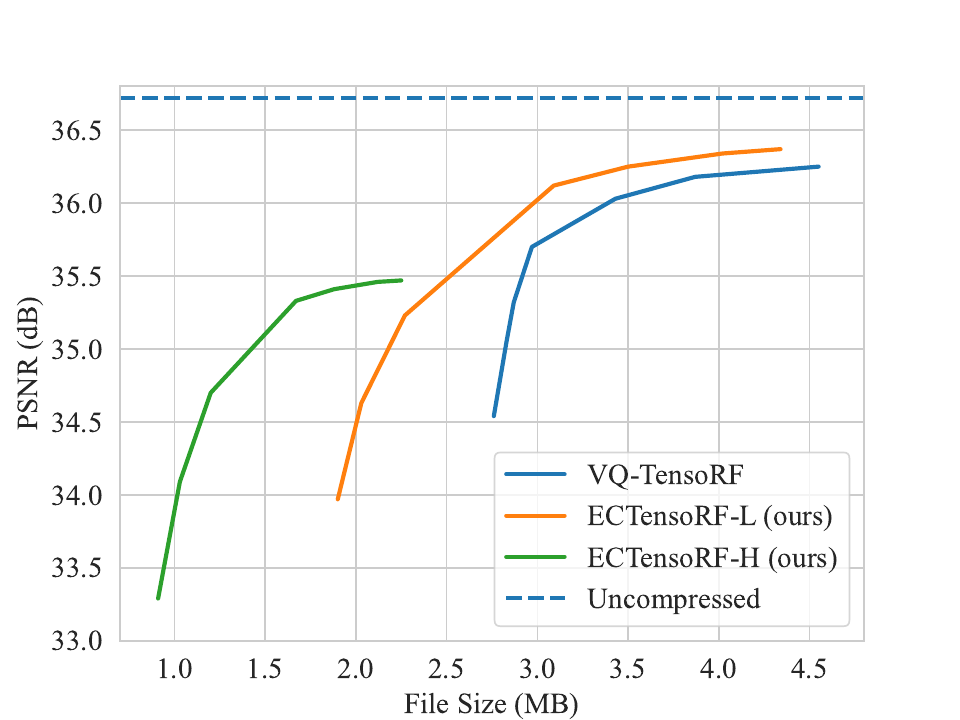}
\includegraphics[width=0.48\textwidth]{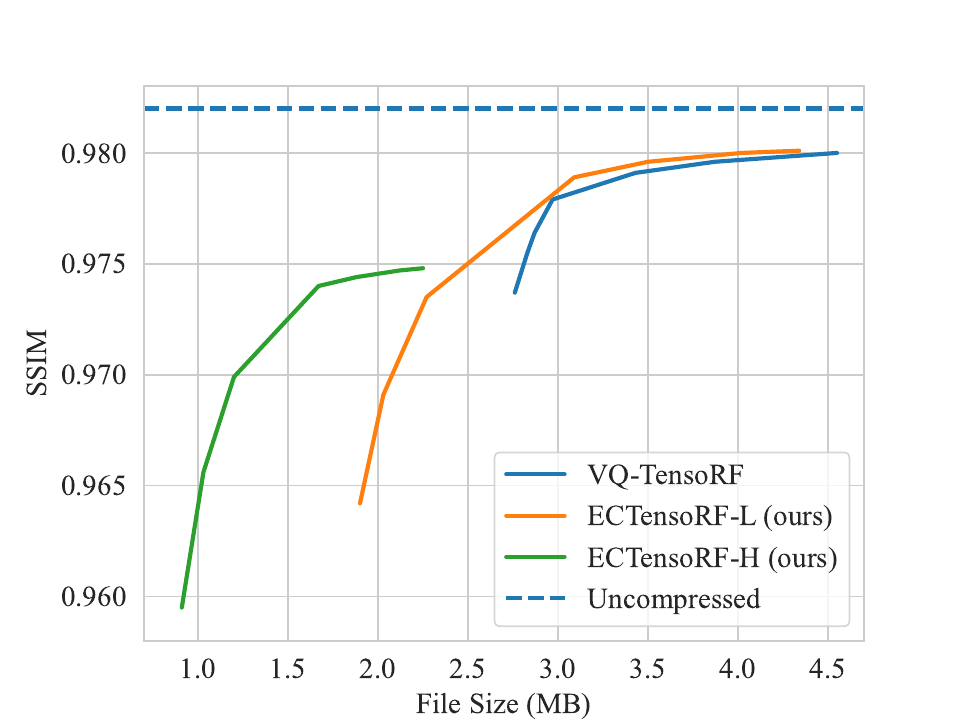}
\caption{Comparison on Synthetic-NSVF dataset.}
\label{fig:rd_psnr_nsvf}
\end{center}
\end{figure*}

\begin{figure*}[h]
\begin{center}
\includegraphics[width=0.48\textwidth]{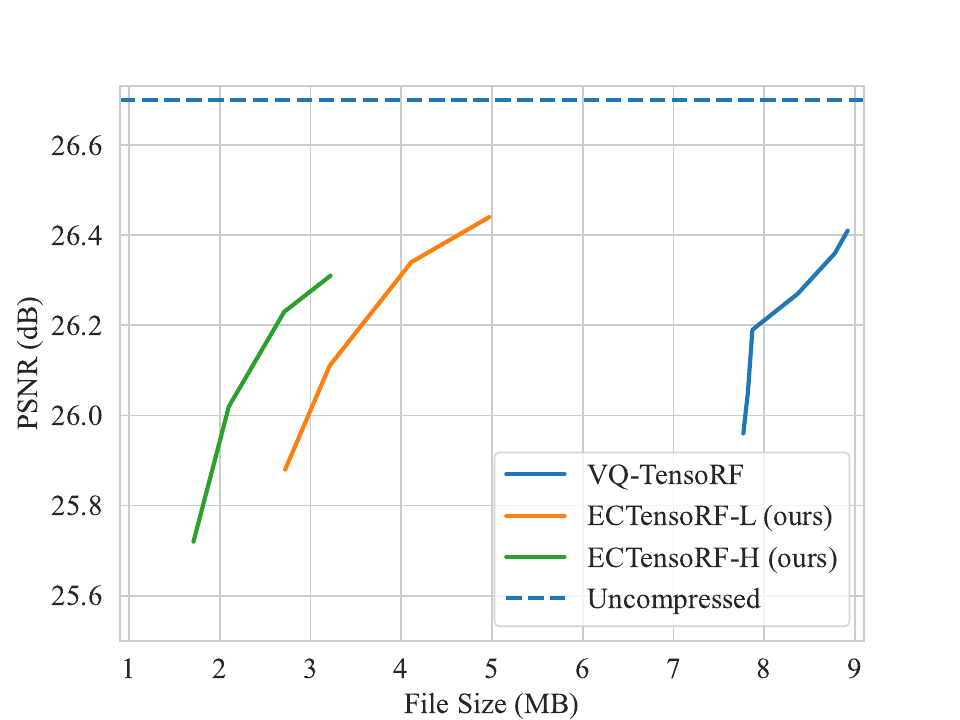}
\includegraphics[width=0.48\textwidth]{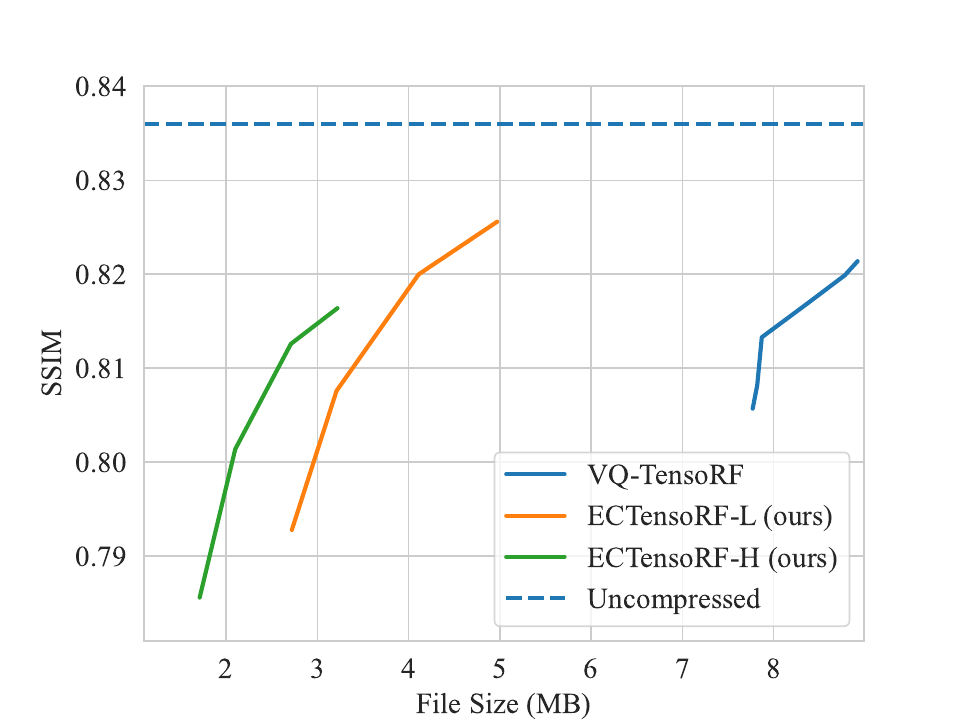}
\caption{Comparison on LLFF dataset.}
\label{fig:rd_psnr_llff}
\end{center}
\end{figure*}

\begin{figure*}[h]
\begin{center}
\includegraphics[width=0.48\textwidth]{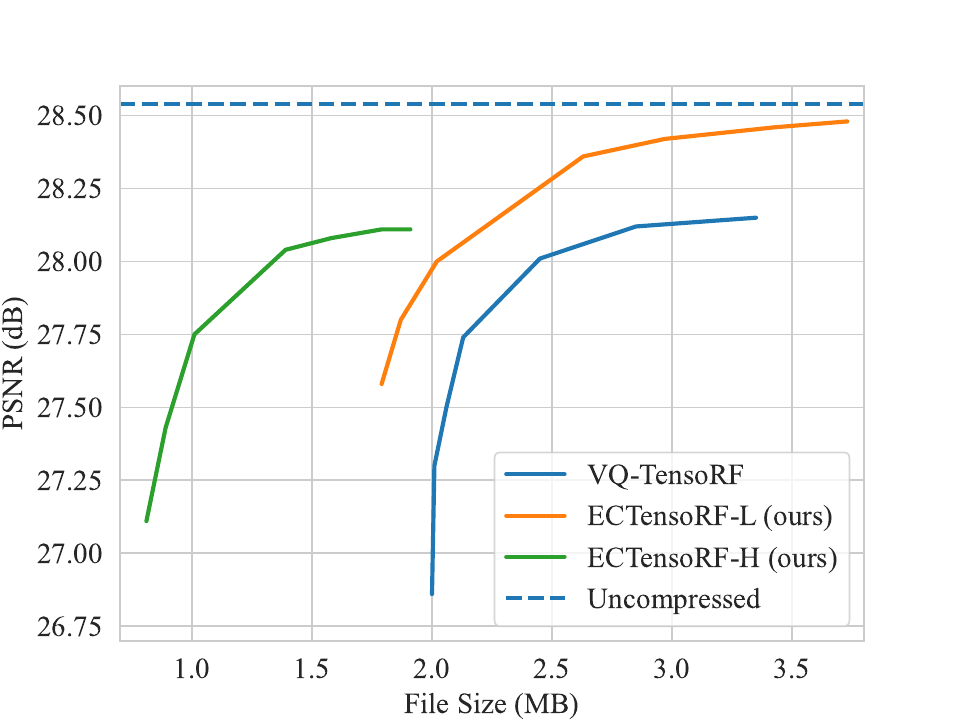}
\includegraphics[width=0.48\textwidth]{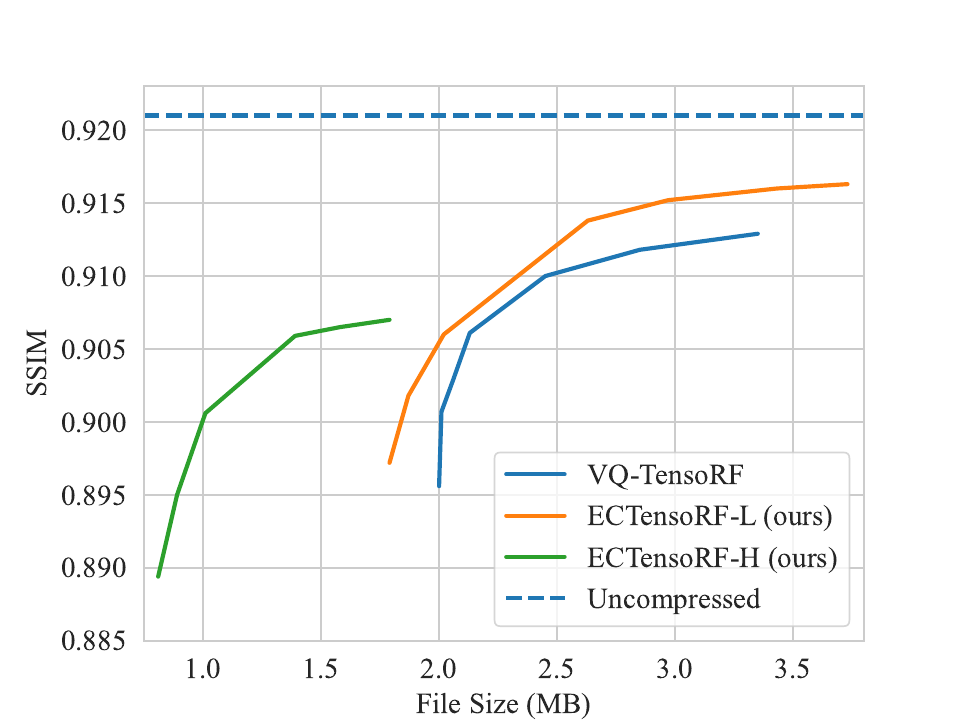}
\caption{Comparison on Tanks and Temples dataset.}
\label{fig:rd_psnr_tt}
\end{center}
\end{figure*}

We further compare the rate-distortion curves of ECTensoRF and the baseline VQ-TensoRF on the Synthetic-NSVF, LLFF and Tanks\&Temples datasets in Figure \ref{fig:rd_psnr_nsvf}, \ref{fig:rd_psnr_llff} and \ref{fig:rd_psnr_tt}.

\subsubsection{Additional experiments}
\paragraph{Latent initialization.} We compare the performance of Gaussian initialization and Zero initialization of the latents code. The results are shown in Table \ref{table:init}.

\begin{table}[H]
\centering
\caption{Comparison of Zero Initialization vs. Gaussian Initialization}
\label{table:init}
\begin{tabular}{|c|c|c|c|c|}
\hline
\textbf{$\lambda$ Values} & \multicolumn{2}{c|}{\textbf{Zero Initialization}} & \multicolumn{2}{c|}{\textbf{Gaussian Initialization}} \\ \hline
                           & \textbf{PSNR (dB)}        & \textbf{Size (MB)}      & \textbf{PSNR (dB)}           & \textbf{Size (MB)}         \\ \hline
2e-2                       & 31.31                     & 1.86                    & 31.13                        & 1.85                       \\ \hline
1e-2                       & 31.80                     & 1.99                    & 31.75                        & 1.99                       \\ \hline
5e-3                       & 32.25                     & 2.20                    & 32.26                        & 2.20                       \\ \hline
1e-3                       & 32.83                     & 3.00                    & 32.83                        & 3.01                       \\ \hline
5e-4                       & 32.93                     & 3.40                    & 32.92                        & 3.42                       \\ \hline
2e-4                       & 32.98                     & 3.92                    & 32.98                        & 3.94                       \\ \hline
1e-4                       & 33.00                     & 4.24                    & 32.99                        & 4.26                       \\ \hline
\end{tabular}
\end{table}

\paragraph{End-to-end training.} We conduct experiments to compare the performance of our two-stage training (by first using a pre-trained TensoRF model, and train the compression model) and a single stage (by training the compression model from scratch). We show the results in Table \ref{table:e2e}.

\begin{table}[H]
\centering
\caption{Comparison of End-to-End Training vs. Two Stages Training}
\label{table:e2e}
\begin{tabular}{|c|c|c|c|c|}
\hline
\textbf{$\lambda$ Values} & \multicolumn{2}{c|}{\textbf{End-to-End Training}} & \multicolumn{2}{c|}{\textbf{Two Stages Training}} \\ \hline
                           & \textbf{PSNR (dB)}        & \textbf{Size (MB)}     & \textbf{PSNR (dB)}           & \textbf{Size (MB)}    \\ \hline
2e-2                       & 25.86                     & 1.69                   & 31.31                        & 1.86                  \\ \hline
1e-2                       & 28.30                     & 1.71                   & 31.80                        & 1.99                  \\ \hline
5e-3                       & 30.05                     & 1.81                   & 32.25                        & 2.20                  \\ \hline
1e-3                       & 31.29                     & 2.39                   & 32.83                        & 3.00                  \\ \hline
5e-4                       & 31.53                     & 2.57                   & 32.93                        & 3.40                  \\ \hline
2e-4                       & 31.86                     & 2.69                   & 32.98                        & 3.92                  \\ \hline
1e-4                       & 31.80                     & 2.95                   & 33.00                        & 4.24                  \\ \hline
\end{tabular}
\end{table}

\paragraph{Hyperprior model.} We also perform experiments using a version of hyperprior model \citep{balle2018variational} with our masking mechanism. More specifically, we apply masking on both the hyper-latents and latents. Both type of latents are directly optimized without using amortized inference. The hyper decoder has two transposed convolutional layers with SELU activation. We show the result on NeRF-Synthetic on Table \ref{table:hyper}.

\begin{table}[H]
\centering
\caption{Comparison of ECTensorF-L with and without Hyperprior}
\label{table:hyper}
\begin{tabular}{|c|c|c|c|c|}
\hline
\textbf{$\lambda$ Values} & \multicolumn{2}{c|}{\textbf{ECTensorF-L + Hyperprior}} & \multicolumn{2}{c|}{\textbf{ECTensorF-L}} \\ \hline
                           & \textbf{PSNR (dB)}        & \textbf{Size (MB)}          & \textbf{PSNR (dB)}         & \textbf{Size (MB)}  \\ \hline
2e-2                       & 31.31                     & 1.92                        & 31.31                      & 1.86                \\ \hline
1e-2                       & 31.92                     & 2.04                        & 31.80                      & 1.99                \\ \hline
5e-3                       & 32.35                     & 2.25                        & 32.25                      & 2.20                \\ \hline
1e-3                       & 32.85                     & 2.97                        & 32.83                      & 3.00                \\ \hline
5e-4                       & 32.93                     & 3.32                        & 32.93                      & 3.40                \\ \hline
2e-4                       & 32.98                     & 3.72                        & 32.98                      & 3.92                \\ \hline
1e-4                       & 33.00                     & 3.95                        & 33.00                      & 4.24                \\ \hline
\end{tabular}
\end{table}

At lower bit rates, the hyperprior is slightly worse than the ECTensoRF-L baseline because of the irreducible cost to transmit the hyper decoder and hyper entropy model. At higher bit rates, the compression performance with the hyperprior method is better than using only a single entropy model, which aligns with prior observations in image compression \citep{balle2018variational}.

\paragraph{Preliminary results for Factor Fields.} We show the potential of applying our method to other grid-based NeRF architectures. We choose Factor Fields \citep{chen2023factor} to experiment with. We show the result of our method for Factor Fields in Table \ref{table:factor}. Note that for Factor Fields, we compress the basis 4D tensors and do not compress the coefficient 4D tensors.

\begin{table}[H]
\centering
\caption{Factor Fields experiments}
\label{table:factor}
\begin{tabular}{|c|c|c|}
\hline
\textbf{$\lambda$ Values} & \textbf{PSNR (dB)} & \textbf{Rate (MB)} \\ \hline
1e-3                       & 26.19              & 1.12               \\ \hline
1e-4                       & 29.67              & 1.23               \\ \hline
1e-5                       & 31.35              & 1.82               \\ \hline
Uncompressed               & 33.09              & 18.89              \\ \hline
\end{tabular}
\end{table}
 
\subsubsection{More Qualitative Results}

\begin{figure*}[h]
    \centering
    \includegraphics[width=1.\textwidth]{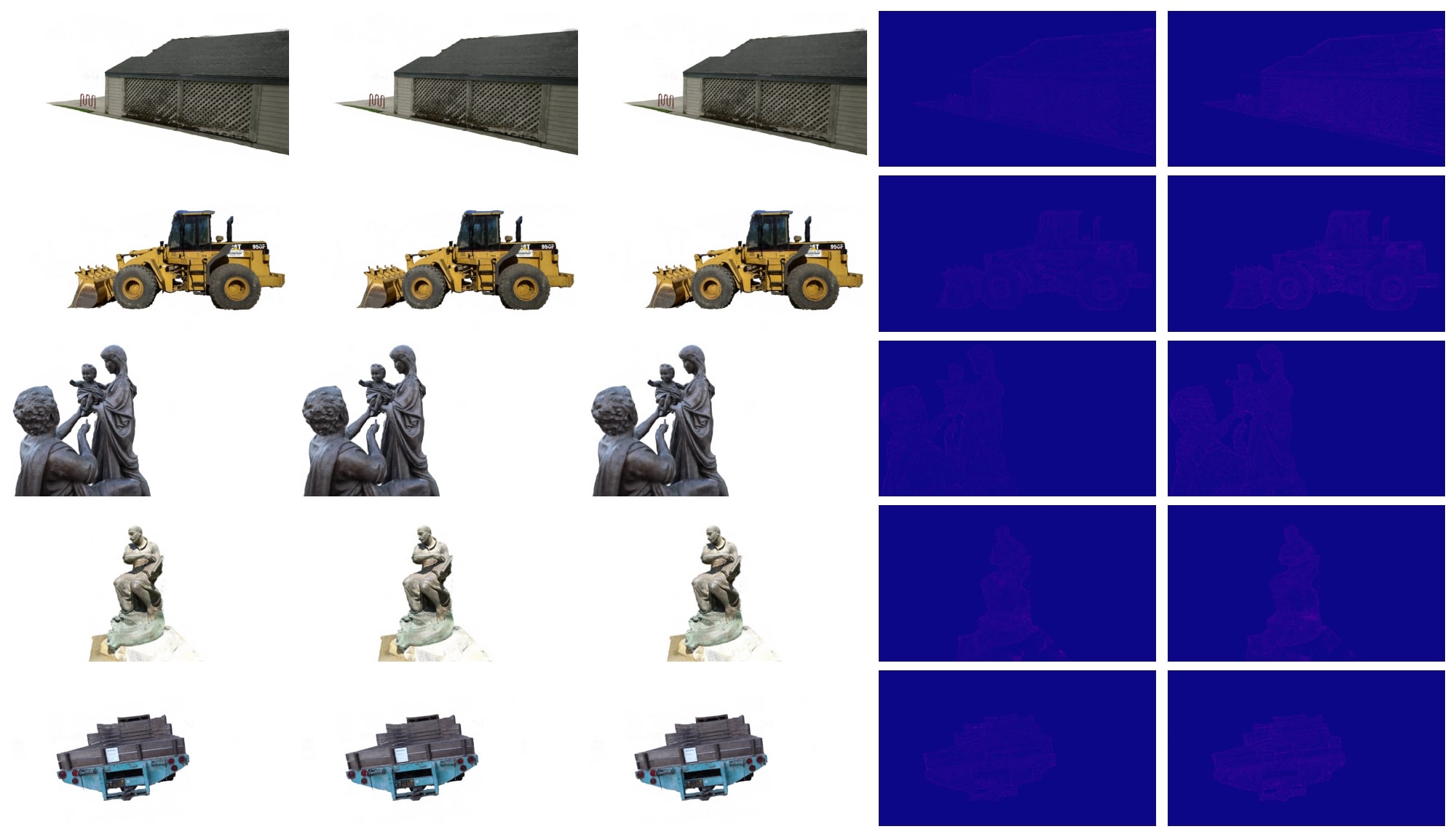}
    \vskip 0.1cm
    \begin{tabu} to 1.\textwidth {X[1,c]X[1,c]X[1,c]X[1,c]X[1,c]}
    \end{tabu}
    \caption{Qualitative results on Tanks and Temples dataset. From left to right: TensoRF, ECTensoRF-L, ECTensoRF-H, ECTensoRF-L difference and ECTensoRF-H difference.}
    \label{fig:q_tt}
\end{figure*}

\begin{figure*}[h]
    \centering
    \includegraphics[width=0.75\textwidth]{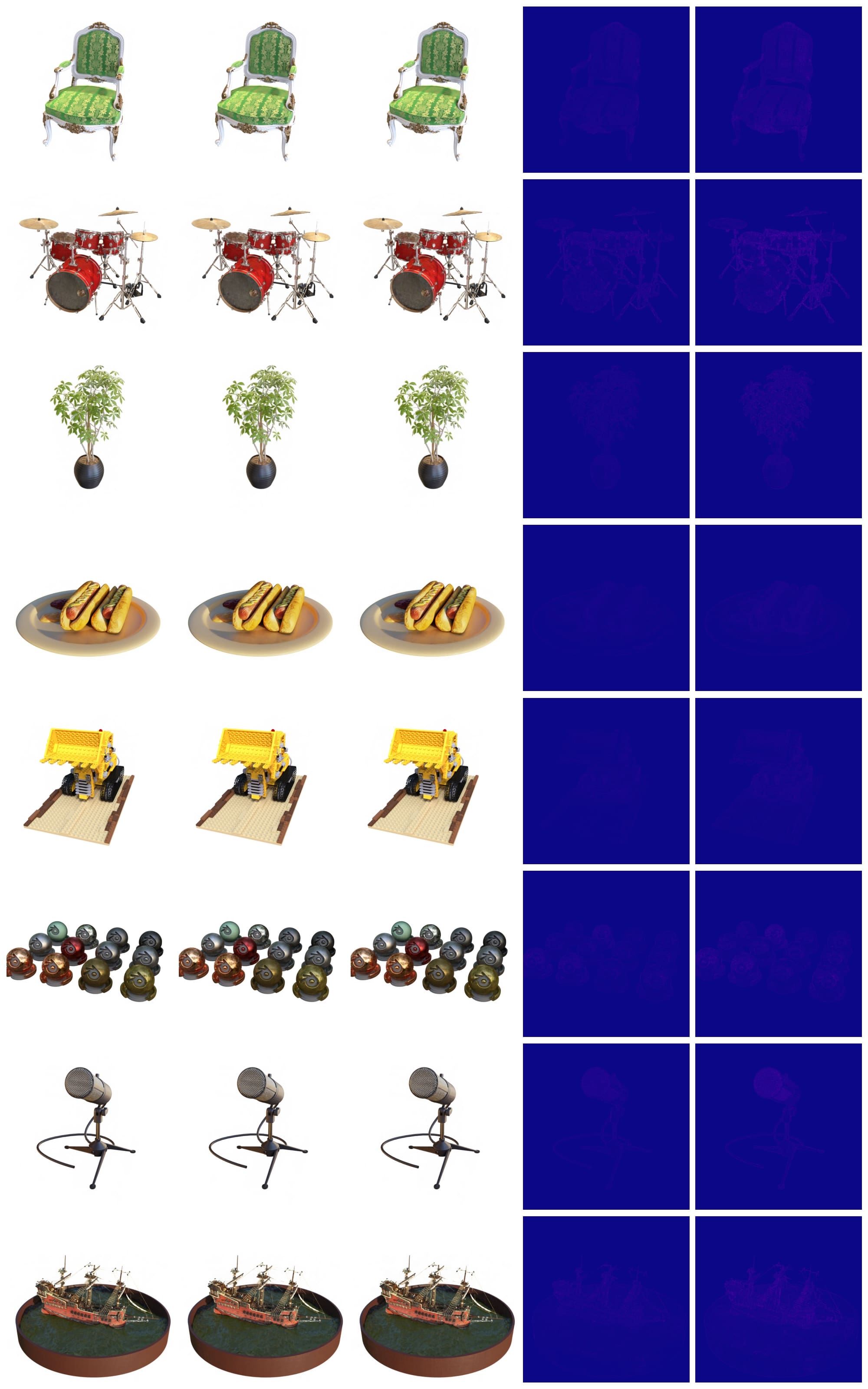}
    \vskip 0.1cm
    \begin{tabu} to 1.\textwidth {X[1,c]X[1,c]X[1,c]X[1,c]X[1,c]}
    \end{tabu}
    \caption{Qualitative results on Synthetic-NeRF dataset. From left to right: TensoRF, ECTensoRF-L, ECTensoRF-H, ECTensoRF-L difference and ECTensoRF-H difference.}
    \label{fig:q_nerf}
\end{figure*}

\begin{figure*}[h]
    \centering
    \includegraphics[width=0.75\textwidth]{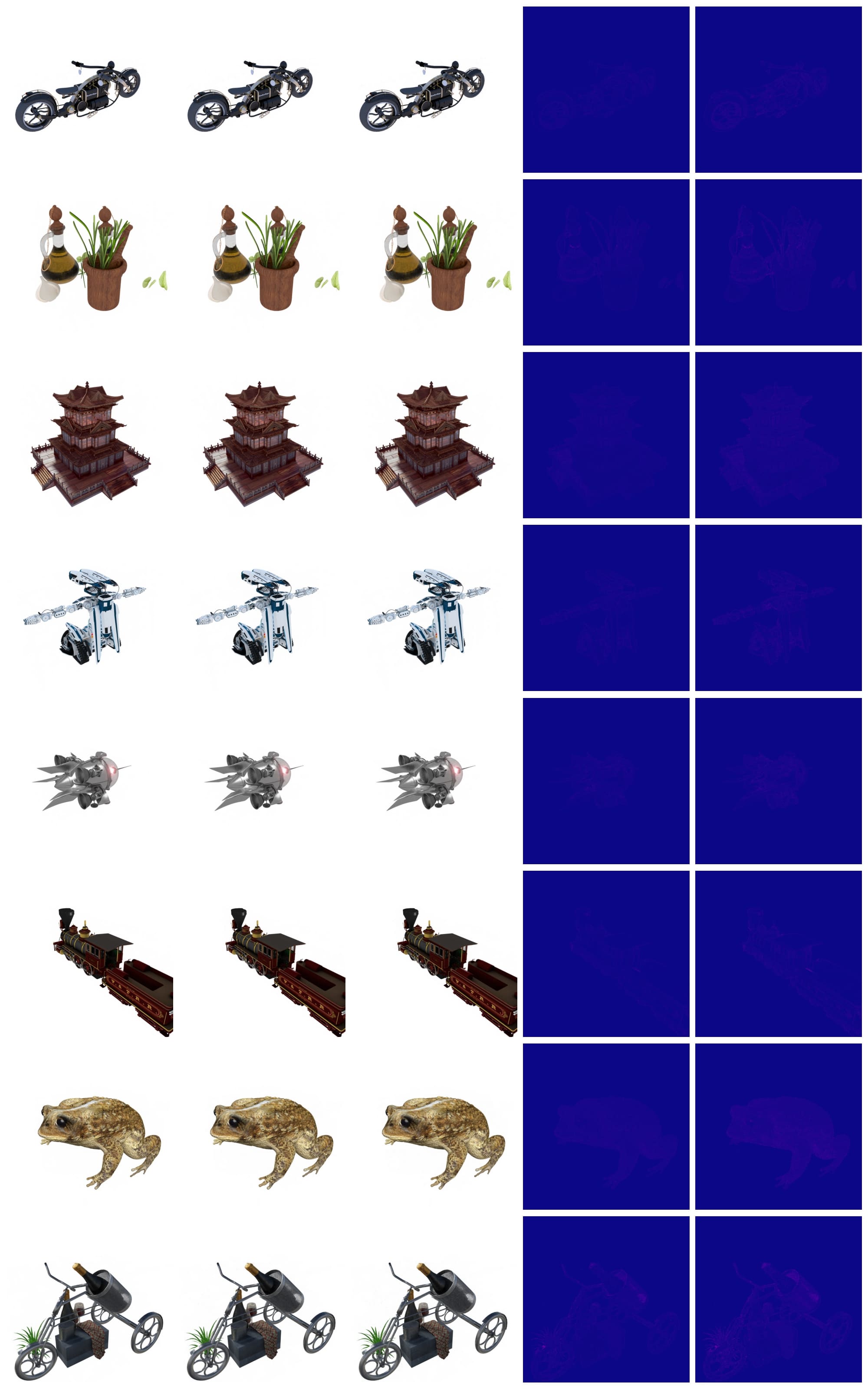}
    \vskip 0.1cm
    \begin{tabu} to 1.\textwidth {X[1,c]X[1,c]X[1,c]X[1,c]X[1,c]}
    \end{tabu}
    \caption{Qualitative results on Synthetic-NSVF dataset. From left to right: TensoRF, ECTensoRF-L, ECTensoRF-H, ECTensoRF-L difference and ECTensoRF-H difference.}
    \label{fig:q_nsvf}
\end{figure*}

\begin{figure*}[h]
    \centering
    \includegraphics[width=1.\textwidth]{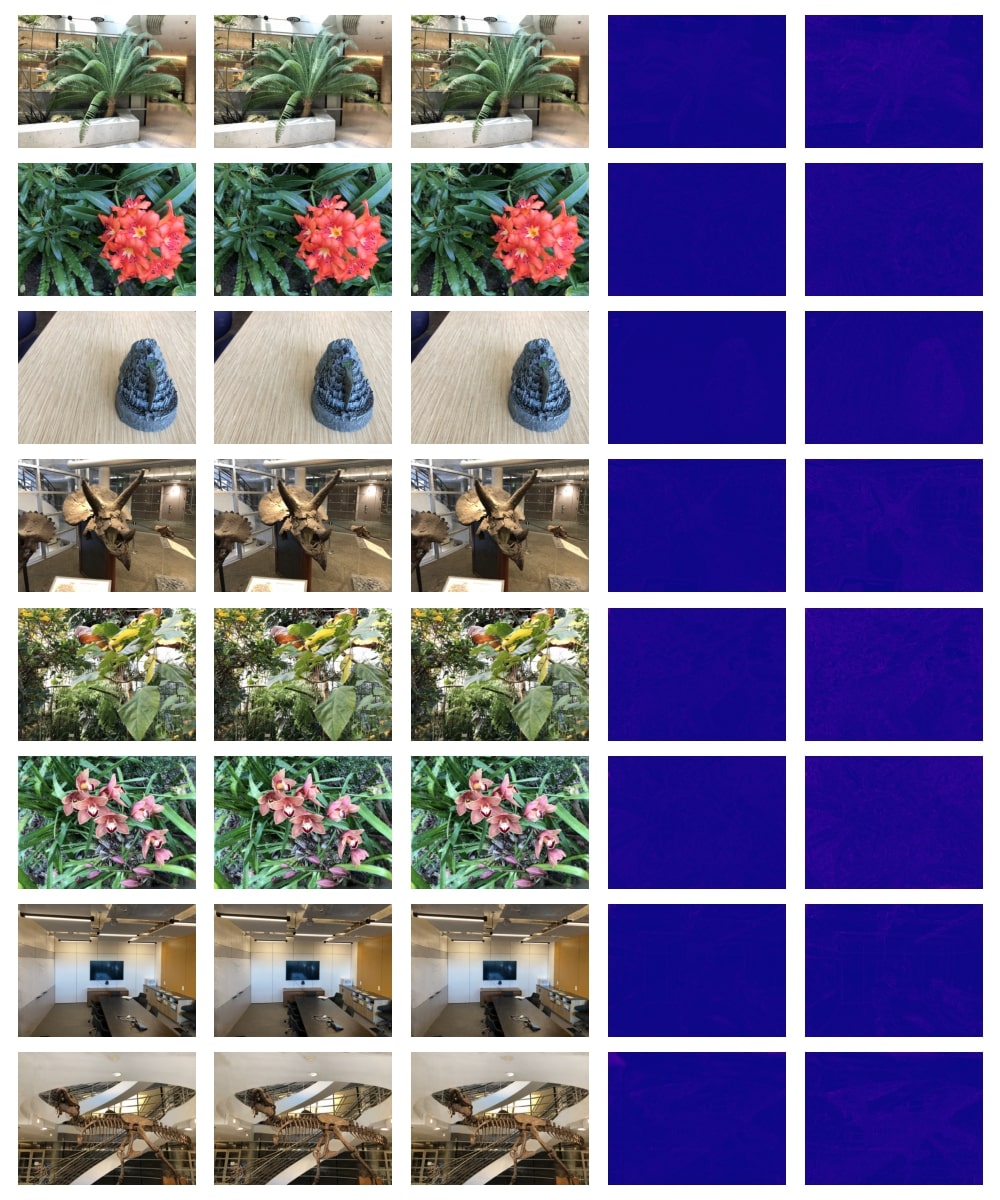}
    \vskip 0.1cm
    \begin{tabu} to 1.\textwidth {X[1,c]X[1,c]X[1,c]X[1,c]X[1,c]}
    \end{tabu}
    \caption{Qualitative results on LLFF dataset. From left to right: TensoRF, ECTensoRF-L, ECTensoRF-H, ECTensoRF-L difference and ECTensoRF-H difference.}
    \label{fig:q_llff}
\end{figure*}

We show qualitative results on all scenes from Synthetic-NeRF, Synthetic-NSVF, LLFF and Tanks\&Temples datasets in Figure \ref{fig:q_tt}, \ref{fig:q_nerf}, \ref{fig:q_nsvf} and \ref{fig:q_llff}.

\end{document}